\documentclass{article}

\usepackage[T1]{fontenc}
\usepackage{microtype}
\usepackage{graphicx}
\usepackage{subcaption}
\usepackage{booktabs}
\usepackage{tabularx}
\usepackage{hyperref}
\usepackage{multirow} 
\usepackage{longtable}
\usepackage{placeins}
\usepackage{footmisc}

\usepackage[accepted]{icml2026}[accepted]

\usepackage{amsmath}
\usepackage{amssymb}
\usepackage{xcolor}
\usepackage{listings}
\usepackage{tcolorbox}
\tcbuselibrary{breakable, skins}
\usepackage{enumitem}
\usepackage{array}
\usepackage{wrapfig}
\usepackage[capitalize,noabbrev]{cleveref}
\lstdefinestyle{promptstyle}{
  basicstyle=\small\ttfamily,
  breaklines=true,
  breakatwhitespace=true,
  breakindent=0pt,
  columns=fullflexible,
  frame=single,
  backgroundcolor=\color{gray!10},
  rulecolor=\color{gray!40},
  xleftmargin=1em,
  xrightmargin=1em,
  extendedchars=true,
  inputencoding=utf8,
  literate=%
    {’}{{\textquoteright}}1
    {‘}{{\textquoteleft}}1
    {“}{{\textquotedblleft}}1
    {”}{{\textquotedblright}}1
    {—}{{---}}1
    {–}{{--}}1
    {…}{{\ldots}}1
    {→}{{$\rightarrow$}}1
    {←}{{$\leftarrow$}}1
    {↔}{{$\leftrightarrow$}}1
    {≥}{{$\geq$}}1
    {≤}{{$\leq$}}1
    {≠}{{$\neq$}}1
    {×}{{$\times$}}1
    {±}{{$\pm$}}1,
}

\newcommand{\papertitle}{LLMs Can Annotate Attribution Graphs}

\newcommand{\codeurl}{\url{https://github.com/maxh119Z/circuit-tracer-automation}}

\newcommand{\dataurl}{\url{https://huggingface.co/datasets/circuit-tracer-automation/pipeline_automation}}

\newcommand{\circuittracer}{circuit-tracer}
 
\definecolor{lightgray}{gray}{0.92}
\definecolor{todored}{RGB}{200,30,30}

\newtcolorbox{promptbox}[1][]{
  enhanced, breakable,
  colback=gray!5, colframe=gray!50, fonttitle=\bfseries,
  left=2mm, right=2mm, top=1mm, bottom=1mm,
  #1
}
 
\icmltitlerunning{\papertitle}

\begin{document}

\twocolumn[
  \icmltitle{\papertitle}

  \icmlsetsymbol{equal}{*}

  \begin{icmlauthorlist}
    \icmlauthor{Ameen Patel}{equal}
    \icmlauthor{Max Zhang}{equal}
    \icmlauthor{Nathan Hu}{aff1}
  \end{icmlauthorlist}

  \icmlaffiliation{aff1}{Stanford University}

  \icmlcorrespondingauthor{Nathan Hu}{nathu@cs.stanford.edu}

  \icmlkeywords{Machine Learning, ICML}

  \vskip 0.3in
]

\printAffiliationsAndNotice{\icmlEqualContribution}

\begin{abstract}
Circuit tracing is an exciting technique for revealing the internal computation of language models, but it requires a time-intensive manual step of grouping individual features or MLP neurons into supernodes. We present a simple pipeline for automating this step: directly presenting feature descriptions to a language model that groups them into supernodes. Using automated interpretability metrics, we confirm that supernodes generated by our pipeline are as interpretable as those generated by human annotators. On a two-hop Capitals task, our pipeline recovers a supernode corresponding to the intermediate hop in 97 of 100 prompts. Finally, we present a simple proof of concept using our pipeline for open-ended exploration, where we automatically annotate 1000 attribution graphs from Wikipedia prompt completions and then use an LLM judge to flag interesting graphs worth human review. We hope this work demonstrates that even simple automation can produce meaningful attribution graph annotations, motivating further work on automated circuit tracing. Code\footnote{\codeurl\label{fn:code}} and data\footnote{\dataurl} are provided.

\end{abstract}

\section{Introduction}
\label{sec:introduction}

Recent work in mechanistic interpretability has produced techniques for revealing the internal computation graphs of language models~\citep{lindsey2025landscape}. These \textit{circuit tracing} techniques produce attribution graphs whose nodes are individual model components, such as MLP neurons~\citep{arora2026languagemodelcircuitssparse} or SAE features~\citep{ameisen2025circuit, lindsey2025biology, marks2025sparsefeaturecircuitsdiscovering}. These nodes can be interpreted via their top-activating text examples. However, circuit tracing by default requires a final manual annotation step in which nodes are grouped into \textit{supernodes}: sets of neurons or features corresponding to similar concepts. This manual step is labor-intensive: fully annotating an attribution graph can take one to two researcher-hours~\citep{lindsey2025biology}.

Recent work has begun to address this time-intensive bottleneck. \citet{arora2026adagautomaticallydescribingattribution} group MLP neurons into supernodes by constructing and clustering attribution profiles for each neuron. \citet{birardi2025probeprompting} groups features into supernodes based on their activations across a set of LLM-generated probe prompts related to the original query. Our work similarly aims to automate the supernode grouping step. We treat circuit tracing as a tool for hypothesis generation\footnote{For example, \citet{lindsey2025biology} surfaced evidence of planning circuits in poetry through attribution graphs and steering experiments. However, fully characterizing such planning is the subject of dedicated follow-up work~\citep{nainani2025detectingcharacterizingplanninglanguage, maar2026whatsplanmetricsimplicit}.}, and ask: \textit{what is the simplest way to annotate attribution graphs so that they yield insight into a model's internal computation?}

We present a simple yet effective pipeline for annotating attribution graphs. For each feature in the pruned attribution graph, we first generate a description from its top-activating exemplars, following standard autointerp~\citep{bills2023language, choi2024automatic, paulo2025automaticallyinterpretingmillionsfeatures}. We then use a language model (GPT-5 mini~\citep{singh2026openaigpt5card}) to group these descriptions into supernodes. Our pipeline costs roughly 3--6 cents on heavily pruned graphs ($\sim$80 nodes) and 61 cents on denser graphs ($\sim$585 nodes).

\begin{figure*}[t]
  \centering
  \includegraphics[width=0.47\linewidth]{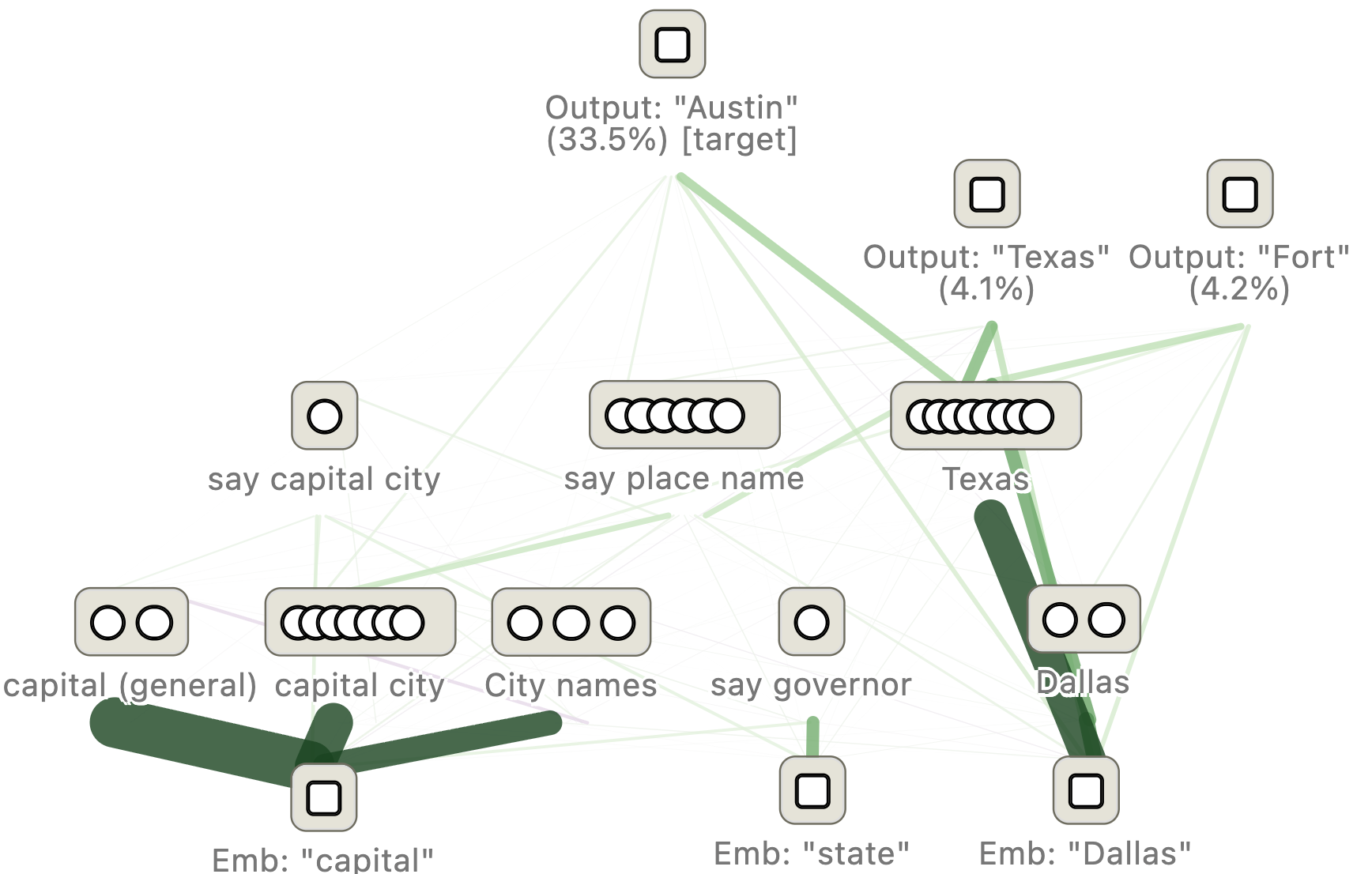}\hfill
  \includegraphics[width=0.49\linewidth]{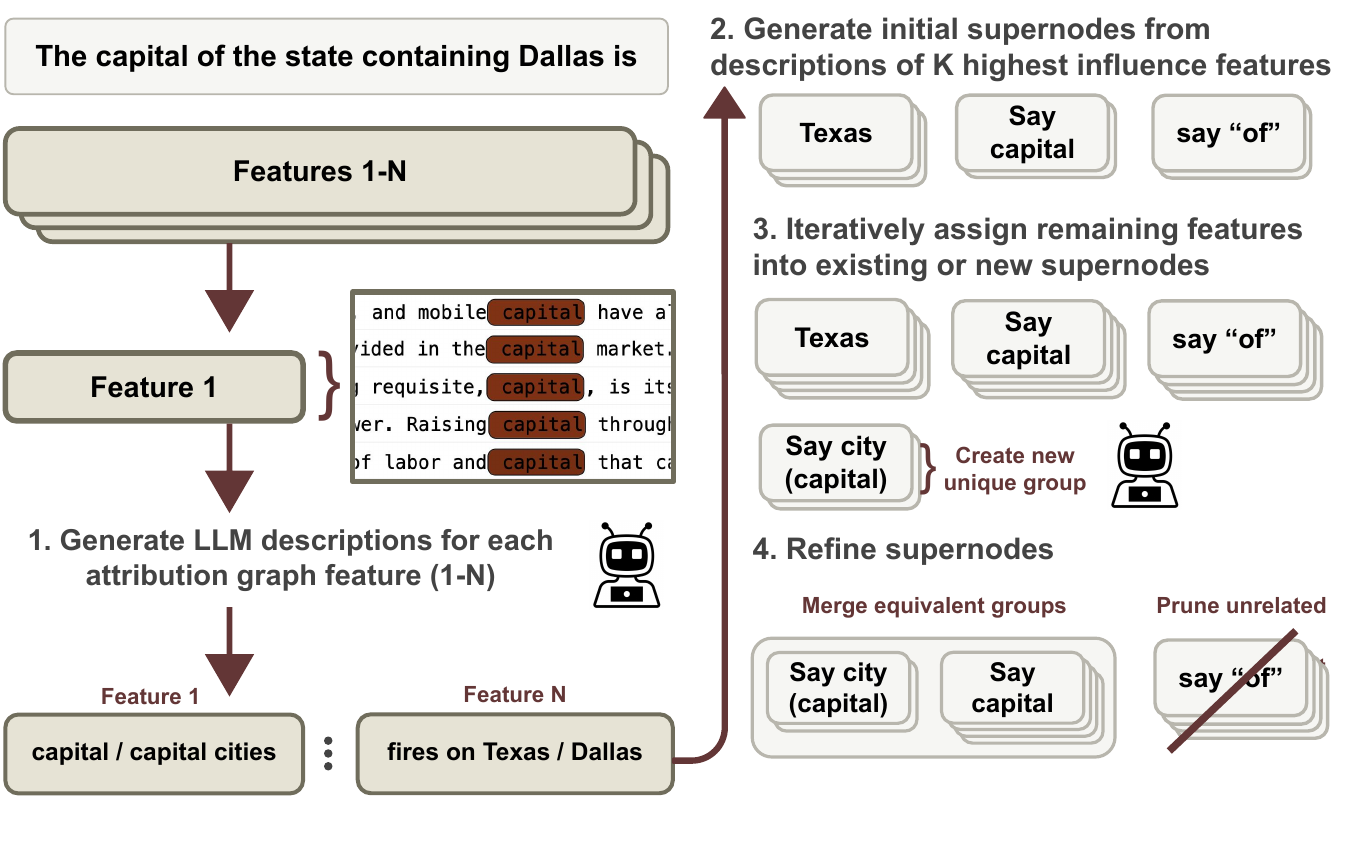}
  \caption{\textbf{Method Overview.} Prompt: ``The capital of the state containing Dallas is'' $\to$ \emph{Austin} at influence threshold 0.40 (42 features). \textbf{Left}: a sample attribution graph annotated by our pipeline. \textbf{Right}: An overview of our LLM supernode grouping process.} 
  \label{fig:method_overview}
\end{figure*}

Our contributions are as follows:
\begin{enumerate}
    \item We present a simple, low-cost pipeline to annotate attribution graphs using an LLM.
    \item We verify that supernodes generated by our pipeline are as interpretable as those from human annotation, according to automated interpretability metrics.
    \item We apply our pipeline to the well-studied two-hop Capitals task and find that it recovers a supernode corresponding to the intermediate hop in 97 of 100 prompts.
    \item Lastly, we present an initial proof of concept for open-ended exploration of attribution graphs: we automatically annotate 1000 Wikipedia prompts and filter the resulting graphs for interesting or surprising behavior.
\end{enumerate}

\section{Methods}
\label{sec:methods}

All experiments use Gemma-2-2B~\citep{gemmateam2024gemma2improvingopen} with the Gemma Scope single-layer transcoders~\citep{lieberum2024gemmascopeopensparse}. We generate attribution graphs through \circuittracer{}~\citep{circuit-tracer} using default hyperparameters. The only hyperparameter we vary is the cumulative node-influence pruning threshold: 0.7 in \Cref{sec:group_validation} and 0.4 in \Cref{sec:experiments,sec:exploration}.\footnote{The 0.4 threshold retains roughly the 40--80 highest-influence nodes per graph.} All steps in our pipeline use GPT-5 mini~\citep{singh2026openaigpt5card} as the LLM annotator.

\subsection{Feature Descriptions}
\label{sec:pipeline-desc}
 
For each feature in the pruned attribution graph, we first use an LLM to generate a feature description from activating text examples. This is the standard autointerp pipeline for feature description \citep{bills2023language, choi2024automatic, paulo2025automaticallyinterpretingmillionsfeatures}. We use 5 activating text examples from Neuronpedia~\citep{neuronpedia}. We additionally provide the LLM with logit information about the tokens most promoted and most suppressed by the feature. We take care to tune our instructions for what features can be described as output features (``say X''). The description prompt and sample descriptions are in Appendix~\ref{app:prompt_annotation} and Appendix~\ref{tab:feature-descriptions}.

\subsection{Supernode Grouping}
\label{sec:pipeline-group}
 
We generate supernodes by presenting the feature descriptions from the previous step to an LLM in three phases. At each phase, the LLM also receives the original prompt and the model's top-3 next-token predictions for additional context.

\begin{enumerate}[leftmargin=*, itemsep=1pt, topsep=2pt]
  \item \textbf{Generate supernodes.} Present the 50 highest-influence features and ask the LLM to identify and name meaningful supernodes, grouping those features into them (or leaving them ungrouped if irrelevant).
  \item \textbf{Iteratively assign remaining features.} Each remaining feature is then presented to either be assigned to an existing supernode, placed in a new supernode (rare, reserved for relevant features that don't fit existing groups), or left ungrouped. We present the remaining features descriptions in batches of 50. 
  \item \textbf{Refine supernodes.} A final pass that drops irrelevant supernodes, merges equivalent ones (directly or in the context of the prompt), and reassigns misplaced features.
\end{enumerate}

As with feature descriptions, we include detailed instructions on what makes a useful supernode. Output supernodes can be formed only by merging output features. Grouping granularity is judged relative to the prompt. For example on 2-hop capitals Dallas prompt, surrounding Southern states might merge into one supernode while Texas remains its own group. More heuristically, we also instruct the model to avoid building supernodes for less informative patterns like stop words or punctuation. We provide full prompts for all stages of our pipeline in Appendix~\ref{app:prompts}.

\paragraph{Cost and Runtime.} Cost scales roughly linearly with the number of features in the pruned graph. On heavily pruned Capitals graphs (40--80 features), our pipeline costs around \$0.05 per graph (34K input and 13K output tokens); on denser graphs at the 0.7 threshold ($\sim$585 features), it costs around \$0.60 per graph (728K input and 214K output tokens).

\section{Comparison to Human-Annotated Supernodes}
\label{sec:group_validation}

\begin{table*}[t]
  \centering
  \small
  \begin{tabular*}{\textwidth}{@{\extracolsep{\fill}} lccccc}
    \toprule
    Condition & \shortstack{Feature\\Detection} & \shortstack{Text\\Detection} & \shortstack{Features\\Grouped} & \# Groups & \shortstack{Feat./\\group} \\
    \midrule
    Human-annotated  & 66.6\% & 81.8\% & 33.7   & 6.6  & 4.91  \\
    Ours (top 50)    & 80.8\% & 81.7\% & 26.9   & 7.1  & 3.98  \\
    Ours (top 100)   & 83.1\% & 82.1\% & 48.5   & 7.1  & 7.08  \\
    Ours (top 150)   & 82.8\% & 84.2\% & 68.1   & 7.5  & 9.43  \\
    Ours (top 200)   & 81.9\% & 81.7\% & 88.5   & 8.1  & 11.62 \\
    Ours (full)      & 93.4\% & 83.2\% & 207.5  & 8.3  & 26.42 \\
    Ours (full, no refine) & 92.3\% & 82.1\% & 250.1  & 11.1 & 24.30 \\
    \midrule
    Random Grouping  & 22.1\% & 54.7\% & --     & --   & --    \\
    Chance           & 10\%   & 50\%   & --     & --   & --    \\
    \bottomrule
  \end{tabular*}
  
  \caption{\textbf{Autointerp scores vs.\ human baseline.} On 15 Neuronpedia graphs from \citet{circuit-tracer}, our supernodes match or slightly exceed human-annotated ones on both automated interpretability metrics. Supernode sizes of human-annotated groups are most comparable to those from running our method on the top 50 or 100 features.}
  \label{tab:autointerp_vs_human}
\end{table*}

We compare the interpretability of supernodes generated by our pipeline to those produced by human annotation, using the 15 human-annotated reference graphs from \circuittracer{}~\citep{circuit-tracer} on a variety of Gemma-2-2B prompts. We adapt two automated interpretability metrics to the supernode setting. For any supernode (whether from our pipeline, human annotation, or a random baseline), we generate a description in two steps: (1) describe each member feature using an LLM, as in \Cref{sec:pipeline-desc}; (2) feed the list of feature descriptions to a second LLM that produces a single supernode description. We use this supernode description to compute two metrics:

\begin{itemize}[leftmargin=*, itemsep=1pt, topsep=2pt]
  \item \textbf{Feature Detection Score.} Given the supernode description and 10 feature descriptions (1 of which is from the supernode), can an LLM identify the supernode feature?
  \item \textbf{Text Detection Score.} Given the supernode description and 10 text samples (5 from supernode features), can an LLM identify the 5 text samples that activate a supernode feature?
\end{itemize}

Negative features and text examples are sampled from ungrouped features in the same pruned attribution graph. We additionally compare to a random baseline, where supernodes are formed by randomly grouping features in the pruned attribution graph. Our pipeline can group any number of features. We show results applying our pipeline to annotate the top $k$ features for differing values of $k$. Supernode sizes of human-annotated groups are most comparable to those from running our method on the top 50 or 100 features.

\paragraph{Validation Results.} Autointerp scores and supernode statistics are shown in Table~\ref{tab:autointerp_vs_human}. Increasing the number of features shown to the pipeline increases features grouped into supernodes. Most additional features are packing into existing supernodes rather than yielding new supernodes. For all settings, automatically generated supernodes are as interpretable or slightly more so than human-annotated ones on both automated interpretability metrics. Ablating the refine step increases the number of supernodes and slightly decreases autointerp scores. We confirm results hold when varying the LLM judge in Appendix~\ref{validation_models} and report per-feature description scores in Appendix~\ref{app:d1d2}. 

\paragraph{Qualitative Examples.} Appendix~\ref{app:gallery} provides a broad gallery of qualitative examples: side-by-side comparisons with \citet{circuit-tracer} human annotations, plus pipeline annotations on Capitals and Wikipedia graphs, the other datasets explored in this paper.

\section{Capitals: 2-Hop Queries}
\label{sec:experiments}
\label{sec:capitals}
\begin{figure*}[t]
  \centering
  \begin{minipage}[c]{0.30\textwidth}
    \centering
    \small
    \begin{tabular}{lc}
      \toprule
      Outcome & \# \\
      \midrule
      Found by pipeline       & 97 \\
      Present but missed      & 2 \\
      Not in pruned graph     & 1 \\
      \midrule
      Total                   & 100 \\
      \bottomrule
    \end{tabular}
  \end{minipage}\hfill
  \begin{minipage}[c]{0.65\textwidth}
    \centering
    \includegraphics[width=\linewidth]{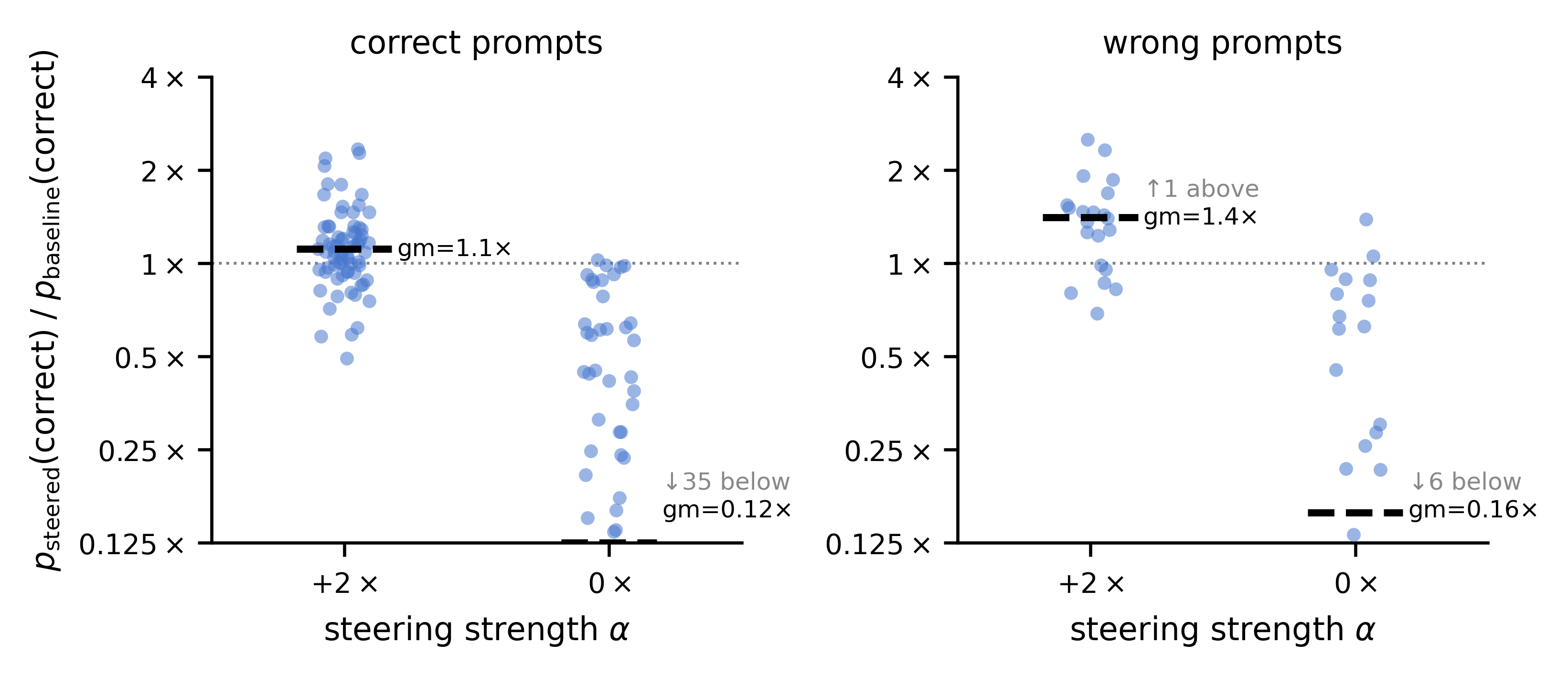}
  \end{minipage}
  \caption{\textbf{Capitals 2-hop results.} \textbf{Left:} Hop recovery outcomes on 100 prompts. Our pipeline recovers a supernode for the intermediate hop in 97 of 100 prompts. We manually evaluated the remaining 3: in 2, an intermediate-hop feature was present but not grouped into a supernode; in 1, no intermediate-hop feature was present in the pruned graph. \textbf{Right:} We confirm via steering that intermediate-hop supernodes are causally important on both correct and incorrect prompts: amplification ($2\times$) increases the probability of the correct answer, suppression ($0\times$) decreases it. Each point is one prompt.}
  \label{fig:capitals_results}
  \label{tab:capitals-hop-outcomes}
  \label{fig:capitals_steering}
\end{figure*}

A well-studied 2-hop circuit~\citep{lindsey2025biology, arora2026languagemodelcircuitssparse, ameisen2025circuit} answers queries such as \textit{``What is the capital of the state containing Dallas?''} by first recalling the state (e.g., \textit{Texas}), then mapping it to its capital. We construct 100 such queries, 50 with each US state as the intermediate hop and 50 with a country. Sample queries are shown in Appendix~\ref{app:datasets}. On each query, we evaluate if our annotation pipeline a supernode corresponding to the intermediate hop. Our pipeline correctly recovers the intermediate-hop supernode in 97 of 100 prompts (\Cref{fig:capitals_results}, left); of the remaining 3, human review found 2 had intermediate-hop features the pipeline missed, while 1 genuinely had no such features in the pruned graph. We then confirm that recovered intermediate-hop supernodes are causally important via constrained patching~\citep{circuit-tracer} (\Cref{fig:capitals_results}, right): suppressing them ($0\times$) reduces the probability of the correct answer while amplifying them ($2\times$) increases it, on prompts where the model is initially correct and incorrect alike. To test that our pipeline generalizes beyond transcoder features, we also rerun it on the model's raw MLP neurons (built with ADAG~\citep{arora2026adagautomaticallydescribingattribution}) on these same 100 prompts, and compare the two decompositions in Appendix~\ref{app:mlp}.

\section{Towards Open-Ended Circuit Exploration}
\label{sec:exploration}

\begin{figure*}[t]
  \centering
  \includegraphics[width=0.44\textwidth]{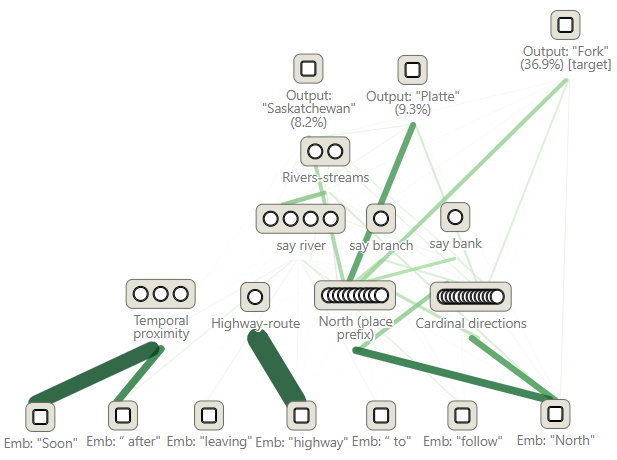}\hfill
  \includegraphics[width=0.55\textwidth]{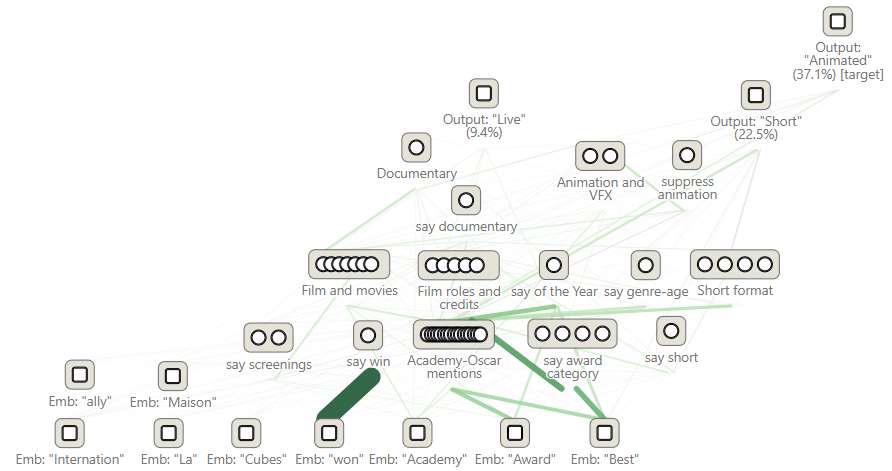}
  \caption{\textbf{Attribution graphs surfaced by open-ended exploration.} \textbf{Left:} ``Soon after leaving the city, the highway begins to follow the North'' $\to$ \emph{Fork}. Despite no mention of water, river-related supernodes are active. Sampling further from Gemma-2-2B yields ``Fork of the Flathead River,'' suggesting these supernodes reflect planning ahead. \textbf{Right:} ``Internationally, La Maison en Petits Cubes won the Academy Award for Best'' $\to$ \emph{Animated}. The graph shows competing categories active in parallel (animation, documentary, movies).}
  \label{fig:exploration}
\end{figure*}

Lastly, we use our pipeline to demonstrate a proof of concept: applying automated annotation to open-ended exploration and identification of meaningful attribution graphs. We first collect 1000 one- to two-sentence Wikipedia snippets, applying heuristic filtering to avoid graphs that predict low-entropy tokens or stop words (full details in Appendix~\ref{app:wiki-construction}). We then use GPT-5.4 to rate each annotated graph along several handcrafted criteria for interestingness (details in Appendix~\ref{app:prompt_judge}). We use these ratings to filter from 1{,}000 graphs down to 10 for manual review, of which we find two interesting graphs we present below. We show these graphs in \Cref{fig:exploration}. In the first graph (left), on the prompt ``Soon after leaving the city, the highway begins to follow the North'' $\to$ \emph{Fork}, many active features relate to rivers and streams despite no mention of water. Sampling further from Gemma-2-2B yields ``Fork of the Flathead River,'' suggesting these river features reflect internal planning in the model. In the second graph (right), on the prompt ``Internationally, La Maison en Petits Cubes won the Academy Award for Best'' $\to$ \emph{Animated}, all of the most likely completions correspond to different Academy Award categories. For each category, there is a corresponding internal supernode, suggesting that the model's internal computation promotes several plausible continuations in parallel.

\section{Discussion}
\label{sec:discussion}

In this work, we present a simple pipeline for automated annotation of attribution graphs, automating a labor-intensive step of circuit tracing. Beyond general interpretability research, automated annotation could enable broad, unsupervised search for interesting attribution graphs; we provide a proof of concept in Section~\ref{sec:exploration}. Additionally, supernode grouping can apply to any feature basis (MLP neurons, single-layer transcoders, or cross-layer transcoders), and an exciting direction for automated annotation is to enable systematic comparison of these different bases for decomposing attribution graphs; we take a first step toward this in Appendix~\ref{app:mlp}, comparing MLP-neuron and transcoder decompositions of the Capitals graphs.

Our work has several limitations. We study a single model and transcoder suite on a limited set of tasks: the 15 reference graphs from \citet{circuit-tracer} and the relatively simple 2-hop Capitals task. We only test our pipeline with a single LLM annotator, GPT-5 mini. Our supernode grouping pipeline is naive: it ignores spatial information (the layer and prompt-token position of each feature), the attribution flow between features, and the backward-from-output search style that human circuit-tracers often use. More generally, human interpretation draws on much more context than activating text examples alone, including the prompt and surrounding features in the graph. For example, in Appendix~\ref{app:math}, we apply our pipeline to arithmetic queries and find that the resulting supernode labels are too coarse to capture the known internal structure. Incorporating these is a clear direction for future work. Despite these many areas for improvement, our method already produces useful annotations of attribution graphs. We hope this work demonstrates that there remains much low-hanging fruit when automating circuit tracing and motivates further work in this direction.

\newpage
\section*{Impact Statement}
This paper presents work that advances the field of mechanistic interpretability. We present a method for automating the human-labor-intensive feature annotation step of circuit tracing, which can accelerate interpretability research more broadly. Research progress in interpretability is inherently dual-use: it can benefit capabilities while also improving our safety and understanding of models. We believe this work is a net positive.

\clearpage
\bibliography{references}
\bibliographystyle{icml2026}

\newpage
\appendix
\onecolumn
\section{Full Prompt Templates}
\label{app:prompts}
 
\subsection{Feature Description Prompt}
\label{app:prompt_annotation}
 
\begin{promptbox}[title=Feature description annotation (default)]
\begin{lstlisting}[style=promptstyle]
You are a mechanistic interpretability researcher. You will be given evidence about a single feature neuron. Your task is to produce a label and brief description for this feature.

You will receive three types of evidence:
1. Overall Prompt Context: the original prompt the model was processing.
2. Input Activations: text excerpts where the neuron activated strongly. The most relevant tokens are delimited by <<<>>>.
3. Global Output Tokens: tokens this neuron tends to push toward or away from in the output.

Use input activations as the primary evidence. Use prompt context only for disambiguation, not as proof by itself. Output tokens can be noisy — only factor them in when they show a clear, consistent pattern. If they are consistent, they likely reveal a lot of information. A tight cluster of specific promoted tokens (e.g. one city, one state) outranks a broader category label — prefer the specific entity.

STYLE: Write in short, direct fragments — not full sentences. Get to the point immediately. No filler, no hedging, no grammatical padding.

FEATURE TYPES — use this to guide your description style:
Features tend to fall into three types. Figure out which one fits, then describe accordingly.

1. Input features — activate on a specific token or category of tokens.
   Describe what they activate on: 'represents X' or just name the pattern directly.
   If they activate on a range of related things, describe the category.

2. Output features — consistently promote a specific next token or category.
   Label as 'say X' when a clear next-token pattern exists. Reference definitions below. Prepositions may fall under this category, where the important words are subsequent to the thing it is referencing.
3. Abstract/middle features — neither cleanly input nor output.
   Describe the context pattern: what kind of text, what situation, what role it plays. These often need the surrounding context of activations, not just the highlighted token.

'SAY X' vs 'X ITSELF':
Features can represent a concept directly, or signal that a concept is about to appear (activating on structural words right before it — prepositions, articles, punctuation).
- Highlighted tokens are content words → SHORT_LABEL is the concept itself.
- Highlighted tokens are structural words setting up content → SHORT_LABEL is 'say [what]'.
- Check what follows the trigger across activations: if a specific concept X (e.g. a proper noun, a method name) CONSISTENTLY appears right after the trigger token, that supports 'say X'. When unclear, prefer naming the concept directly 'X' without the 'say' — 'say X' is a stronger claim and needs consistent evidence and should not be used lightly: be strict about including 'say' in any feature.

PROPER NOUNS:
If a specific name, place, or entity recurs across the activations — even in a minority of them — include it in the SHORT_LABEL or elaboration. Don't collapse to a generic label when a specific one is clearly supported. Beyond highlighted triggers, also consider consistently occurring proper nouns. These are a signal of specificity, not noise. If highlighted tokens vary widely and the feature looks polysemantic, capture the consistently specific entities that recur across excerpts when clear and possible rather than defaulting to a single broad label.

AVOID:
- Linguistic or technical jargon: copula, lemma, morpheme, orthogonal, syntactic, prepositional phrase, noun phrase, etc. Prefer layman's vocabulary and casual tone.
- Broad labels when something more specific is clearly supported.
- Full sentences. Filler (grammar is not required).

OUTPUT FORMAT: SHORT_LABEL — elaboration

- SHORT_LABEL: 1-5 words. Natural graph node name — specific over generic.
- After ' — ': 1-2 tight fragments. Add context, what it promotes, or consistent subpatterns. Skip if the label already says it all.
- Total: 10-35 words.

Return only the formatted line, nothing else.
\end{lstlisting}
\end{promptbox}

\subsection{Supernode Grouping Prompts}
\label{app:prompt_grouping}
\textbf{Design principles.} The grouping prompts encode several principles that shape the resulting supernodes: features weakly related to the prompt or output are sent to UNGROUPED rather than forced into groups; specific labels (especially recurring proper nouns) are preferred over broad categories; promotion features ("say X") are kept separate from concept features ("X" itself), and "suppress X" is kept separate from both; grammatical or structural patterns default to UNGROUPED unless central to the prompt; and supernode names are constrained to layman's vocabulary, under five words, and at most one or two concepts. These are operationalized across the three phases below.

All three phases use GPT-5 mini. Phase~1 clusters the top \texttt{GROUPING\_TOP\_K\_SEED=50} most-influential features; Phase~2 assigns the remaining features in batches of \texttt{GROUPING\_BATCH\_SIZE=50}; Phase~3 refines. Each phase prompt inlines the same \texttt{GROUPING\_PHILOSOPHY} block (shown once below) and a \texttt{SPECIFICITY\_GUIDANCE} block (default variant \texttt{a2}, also shown once below); alternative variants of the bias block are supported in code. To avoid repetition, the per-phase prompt boxes render those shared blocks as \texttt{\{GROUPING\_PHILOSOPHY\}} and \texttt{\{SPECIFICITY\_GUIDANCE\}} placeholders.

Runtime-substituted placeholders in the text below: \texttt{\{prompt\_text\}} is the prompt under analysis; \texttt{\{output\_context\}} is a line of the form ``Model's top predicted outputs: \ldots''; \texttt{\{groups\_context\}} is a JSON dict of current groups and rationales (Phase 2 only); \texttt{\{group\_summary\}} is a breakdown of current groups with member features and descriptions (Phase 3 only); \texttt{\{features\}} is a list of lines of the form ``ID: <feature\_id> $|$ Desc: <description>''.

\begin{promptbox}[title=Shared \texttt{GROUPING\_PHILOSOPHY} block]
\begin{lstlisting}[style=promptstyle]
GOAL: Produce a cohesive attribution graph that highlights the main intent and meaning of the prompt.


RELEVANCE & UNGROUPED:
- Assign to "Ungrouped": weak, isolated, or noisy features; features not meaningfully connected to the main prompt and output semantics.
- Purely grammatical tokens (prepositions, articles, conjunctions, punctuation, copulas) go to "Ungrouped" unless they clearly promote a semantically meaningful role (this is very rare). Sentence structure-level grammar is mostly pointless, unless the prompt or output is about it.
- "say X" groups are ONLY valid when X is a meaningful content word or category (e.g., "say a color", "say a fruit"). Do NOT create "say X" groups when X is a function word, grammatical structure, or syntactic role — e.g., "say 'is'", "say a relative clause", "say 'of'", "say a preposition" are never valid groups. These belong in Ungrouped.
- A valid group should feel connected to at least one other group in the graph, not like an isolated curiosity.
- "Ungrouped" is not a failure.


GRANULARITY & SPECIFICITY:
- Create only groups clearly supported by the data. Prefer the most specific name the evidence supports over broad buckets.
- Preserve meaningful distinctions in abstraction level when relevant to the prompt — don't blindly merge a broad category with a narrower stable subtype.
- Distinctions that explain WHY the model chose one output over another should be preserved.
- The prompt commits to one active sense of every word in it. Features from alternate senses of a word (senses the prompt does not require) may form their own groups — Phase 3 will consolidate them.


SEMANTIC ROLE — "SAY X" vs "X ITSELF" (high-priority rule):
- A feature that promotes a concept is different from a feature that IS the concept. Keep them in separate groups.
  - Highlighted tokens are function/structural words → the feature sets up what follows; name it "say [what]".
  - Highlighted tokens are content words → the feature represents the concept directly; name the concept.
- say tags in descriptions are strong signals — respect them.
- "Say" is for genuine promoting. Do not use it when features directly represent a concept.
- Only use 'say' in a group name when member descriptions themselves use it. Do not add 'say' by inference.
- Surface overlap is never enough reason to merge groups with different semantic roles.


NAMING (STRICT):
- LIMIT: 5 words maximum. No exceptions.
- Natural, simple phrasing. If you need more than 5 words, the name is too specific.
- Before naming a group, read the member descriptions. If a specific named entity (place, person, concept) recurs across them, use that name — do not default to a generic category when the descriptions clearly point to something specific.
- This applies to say-X groups too: "say California" is better than "say a place" when descriptions consistently name California.
- Avoid parentheses and words like "mention", "reference", "entity", "concept", "topic", or "pattern" when a simpler phrase works.
- Prefer layman's vocabulary.


SPLITTING vs MERGING:
- Split when a group mixes semantic roles or abstraction levels.
- Small groups are fine if they are interpretable and prompt-relevant.
\end{lstlisting}
\end{promptbox}

\begin{promptbox}[title=Shared \texttt{SPECIFICITY\_GUIDANCE} block (default variant \texttt{a2})]
\begin{lstlisting}[style=promptstyle]
SPECIFICITY GUIDANCE:
When in doubt between narrower and broader groups, use whichever granularity best explains the model's specific output and prompt. Consider both the prompt and the predicted output together: a distinction is worth keeping only if it is relevant to what was asked and (or) what the model predicted.


BORDERLINE FEATURES: prefer assigning a borderline feature to a plausible existing group over Ungrouped — reserve Ungrouped for features with no meaningful connection to the prompt or output.


MERGE CONSTRAINT: when two same-role groups tell the same part of the story and their separation does not help a reader understand the reasoning differently, merge them. Never merge a say-X group with a concept group — promotion and content always stay separate. Prefer keeping more proper noun specificity when relevant to the prompt and output.


DESCRIPTION-AWARE NAMING: before naming any group, scan the member descriptions for recurring proper nouns or specific named entities. If a specific entity (a place, person, concept) appears consistently across descriptions, use that specific name — do not collapse to a generic category like 'a place' or 'a city' when the descriptions clearly name something specific. Apply the same rule to say-X groups: if descriptions consistently name a specific entity after the trigger, prefer 'say California' over 'say a place'. If features clearly relate to an alternate sense of a key prompt word, name the group with a sense qualifier (e.g. 'X (general)') rather than a domain label that is not applicable given the prompt and output context (e.g. 'economic X' or 'X (music)'). The last few words of a description often carry non-trivial specificity — use them as an additional signal when placing features into groups. Group names must come from member descriptions, not from the prompt and output context. Be faithful to feature descriptions in naming. Treat say-X / X-itself separation as a hard constraint. Every say-X group must contain only promoting features; every concept group must contain only content features. A single misplaced feature is enough to split or reassign.
\end{lstlisting}
\end{promptbox}

\begin{promptbox}[title=Phase 1 (Discover): propose candidate supernode labels]
\begin{lstlisting}[style=promptstyle]
You are an expert AI interpretability researcher analyzing internal representations of a large language model.
Context: The model was given the following prompt: {prompt_text}


{output_context}


Below are the 50 most influential features that activated during this prompt.
Cluster them into meaningful semantic groups ("supernodes").


{GROUPING_PHILOSOPHY}


Additional guidance for this phase:
- A single feature may form its own group only if it reflects a stable, reusable semantic pattern, not a one-off surface detail.
- Prefer names that make the graph easy to read over taxonomically tidy labels.
- Prefer two narrow groups over one vague bucket — Phase 3 later can merge, but cannot recover lost distinctions easily. When a concept is clearly relevant to the prompt or output, err toward creating a group rather than Ungrouped.
- HARD RULE — do NOT create groups for grammatical or structural patterns under any circumstances. Assign those features directly to Ungrouped. This includes: prepositions and locational connectors (say 'of', say 'in', say 'after', say locational preposition), copulas and predicate framing (say 'is', predicate framing, copula, say noun after copula), sentence-completion or next-token patterns (say completion, say next noun), subword or token-prefix fragments, heading markers, where-clause framing, structural/relational patterns derived from words in the prompt itself (containment verbs, prepositional structures, syntactic connectors), typographic or capitalization patterns (title case, capitalized tokens, proper noun formatting), and word-onset or prefix fragments — these describe token shape, not meaning. The test: does this group name a semantic concept, or does it describe a syntactic role or sentence structure? Concept = valid group. Sentence structure = Ungrouped.
- Do not create groups named for prompt format or input structure (e.g. "fact prompt", "fill-in-the-blank") — these describe the wrapper, not the reasoning content. Assign to Ungrouped.


{SPECIFICITY_GUIDANCE}


Features:
{features}
\end{lstlisting}
\end{promptbox}
 
\begin{promptbox}[title=Phase 2 (Assign): batch assignment]
\begin{lstlisting}[style=promptstyle]
You are an expert AI interpretability researcher analyzing internal representations of a large language model.
Context: The model was given the prompt: {prompt_text}


{output_context}


Current groups and rationales:
{groups_context}


{GROUPING_PHILOSOPHY}


Task: Assign each feature below to the best matching existing group.
These are lower-influence features — they rarely promote meaningfully new semantic concepts beyond what Phase 1 already captured. Default to an existing group or "Ungrouped".
Do not force a match: if no group fits clearly, "Ungrouped" is correct.
Only create a new group if the concept is genuinely absent from the existing groups, clearly relevant to the prompt, and specific enough that multiple features would share it — this should be rare.


{SPECIFICITY_GUIDANCE}


Features:
{features}
\end{lstlisting}
\end{promptbox}
 
\begin{promptbox}[title=Phase 3 (Refine): merge and prune]
\begin{lstlisting}[style=promptstyle]
You are an expert AI interpretability researcher reviewing the output of an automated feature grouping pipeline.


Context: The model was given the prompt: {prompt_text}


{output_context}


The pipeline produced {num_groups} groups from {num_features} features. Your job is to clean up the result — rename unclear groups, reassign misplaced features, and drop irrelevant groups as defined below.


{GROUPING_PHILOSOPHY}


Your job is limited to five things only:


1. GRAMMAR KILL: Any group whose name describes a syntactic role, sentence structure, token pattern, word-prefix fragment— move its members to Ungrouped and dissolve it. Examples: "containment verb", "prefix 'ill'", "say location after of", "fill-in-the-blank", "[concept] prefix", "[X] relation", "[X] clause", "location clause". The test: does this name a concept or describe sentence structure / token shape? Structure/shape = dissolve. For borderline say-X groups, judge by X: if X names a concept relevant to the prompt or output reasoning chain, keep the group — the "say" promoting does not make it irrelevant. Exception: if a word is clearly semantic and central to the prompt's reasoning chain, judge by its role in context rather than its word class alone.


2. ALTERNATE SENSE: A word has an alternate sense when it shares a surface form with the relevant concept but means something different given this prompt — this includes any domain (financial, architectural, political, etc.) that the prompt does not require. Example: "notes (music)" or "musical notes" when the prompt asks about note-taking. If alternate-sense groups are present, merge them together into a single fallback group named "[concept] (general)" — do not touch the correct-sense group. Do not split the correct-sense group to create a (general) variant; only create "[concept] (general)" by merging existing alternate-sense groups. If no alternate-sense groups exist, take no action. This applies to genuine alternate senses only — do not use this to merge a specific named group into a broader same-sense group. '[concept] (general)' is strictly for features activating on a genuinely different dictionary definition (e.g. 'bank' as a financial institution vs. a riverbank) — not the same concept in different contexts or positions.


Lastly, suppression-flavored groups ("suppress X", "demote X", "anti-X", "avoid X", "inhibit X") must never be merged into the concept group "X" — they represent opposite causal roles. If multiple suppression variants for the same concept X exist, consolidate them into one group named "suppress X". Always use "suppress X" as the canonical name.


3. RENAME: Are any group names unclear, longer than 5 words, or use jargon? Rename for clarity. A rename must not lose specificity, promote a structural name, or flip a concept group to a say-X group or vice versa. Do not drop intermediate reasoning steps from a group name.


4. REASSIGN: Are any individual features obviously in the wrong group given their description and the prompt? Move them. Only reassign with high confidence.


5. RELEVANCE DROP: If a group's concept has no clear connection to the prompt's reasoning chain or predicted output — it is not a named entity in the prompt or is not specific and interesting in general, not an intermediate reasoning step, and not a framing pattern for the output — drop it (members to Ungrouped). Use the SPECIFICITY GUIDANCE to judge relevance. Exception: keep groups that name a competing value in the same category as the answer (e.g., a wrong language when the prompt asks about a language), and lean towards keeping neighbors or related topics in the same domain (e.g., neighboring states, nearby countries) — these may be informative competing signals, not noise.


SPECIFIC → BROAD PROTECTION: Before any merge or rename, check — is one group semantically more precise than the other (a named entity, specific concept, or something referenced in the prompt or output)? If yes, protect the specific group. "say color" must not collapse into "say appearance"; "say school" must not collapse into "say place name". If the specific group is irrelevant to the reasoning chain, send it to Ungrouped — never collapse into a vaguer group.


Only make changes you are CONFIDENT about. If the grouping looks good, return empty lists for all actions.


{SPECIFICITY_GUIDANCE}


Current grouping:
{group_summary}
\end{lstlisting}
\end{promptbox}

\subsection{Screening Judge Prompt}
\label{app:prompt_judge}

The screening judge is called once per graph with the system prompt below (which lists all dimensions together), followed by the user message template. It scores each graph on five independent dimensions on a 1--10 scale, with a one-sentence justification per score that names the specific supernodes driving the assessment.

\begin{promptbox}[title=Screening judge system prompt]
\small\ttfamily
You are a mechanistic interpretability researcher evaluating attribution graphs from a language model. You score each graph on several dimensions of interestingness, each on a 1--10 scale, and give a short justification for each score.

The dimensions are:
\vspace{0.5em}

1. Hidden intermediate concept / multi-hop reasoning --- Score high if EITHER of these holds (they are two faces of the same thing: internal computation that goes beyond a single surface lookup). (a) HIDDEN INTERMEDIATE CONCEPT --- a supernode represents a concept that is load-bearing for the answer but appears in NEITHER the prompt NOR the predicted output; the model computes it internally as a stepping stone (Paper: implicit `Texas' in Dallas->Austin; the Apollo program behind `space'). Trivial restatements of a prompt/output token do not count. (b) MULTI-HOP REASONING --- the supernodes form a genuine chain (input concept -> one or more intermediate concepts -> output), i.e. composed reasoning steps rather than a single direct association (Paper: multi-step factual reasoning). Medium: one plausible unstated association OR one clear intermediate step beyond lookup. High: a specific, genuinely surprising hidden concept clearly doing real work, OR a multi-link chain where each hop is a distinct, necessary concept.

2. Parallel / competing computation --- The model computes several candidate outputs or competing hypotheses at once, OR combines independent pathways into the answer (e.g. a rough estimate pathway plus an exact-lookup pathway). (Paper: addition's parallel approximate+precise routes; planning multiple candidate words.) Near-synonyms or reformattings of the same token do NOT count. Medium: multiple candidates present but weakly separated. High: clearly distinct competing or complementary pathways.

3. Abstraction beyond surface tokens --- The most influential supernodes operate on abstract, semantic, or relational concepts decoupled from the literal surface tokens --- category-level, language-independent, or role-based features --- rather than echoing specific prompt words. (Paper: language-independent `meaning' features in multilingual circuits.) Medium: some abstraction beyond the tokens. High: core computation is clearly in an abstract concept space, not in the surface strings.

4. Unexpected mechanism (headline) --- Taken as a whole, does the internal computation DIVERGE from what a knowledgeable person would predict just from the prompt and the predicted output? This is the open-ended `reveals computation different from what one might expect' signal. (Paper: chain-of-thought unfaithfulness; mechanisms that contradict the model's own account.) Score LOW when the supernodes are exactly the obvious ones a person would guess. Score HIGH when the route to the answer is structurally different from the naive expectation. Judge the mechanism, not whether the prompt topic is interesting.

5. Other interesting structure --- A catch-all for notable internal structure not covered above. If you notice something genuinely worth a researcher's attention, score it high here and name it specifically.

\vspace{0.5em}
Scoring guidance (use the full 1--10 range):\\
~~1--2 = no evidence of this property\\
~~3--4 = weak or ambiguous hint\\
~~5--6 = plausible but not strongly supported\\
~~7--8 = clear, concrete instance worth noting\\
~~9--10 = textbook example, worth a writeup

Score each dimension independently on its own merits. Do NOT inflate scores because the prompt is interesting; only the supernode structure matters. For each score, provide a one-sentence justification that names the specific supernodes or features driving your assessment.
\end{promptbox}

\begin{promptbox}[title=Screening judge — user message template]
\small\ttfamily
Prompt given to the model: "\{prompt\}"\\
Model's predicted output: "\{predicted\}" --- correct answer is: \{correct\_answer\}\\
Model confidence: \{confidence\} --- \{confidence\_band\}\\
Supernode groups (\{num\_supernode\_groups\} total):\\
\{groups\}

Sample descriptions of the most influential individual nodes (by influence score):\\
\{clerp\_samples\}

Score this graph on each of the five dimensions. Reply in this exact JSON format --- every dimension is required, with both a score and a one-sentence reason:

\{\\
~~"hidden\_or\_multihop": \{"score": <integer 1-10>, "reason": "<one sentence naming the specific supernodes or features>"\},\\
~~"parallel\_or\_competing": \{"score": <integer 1-10>, "reason": "<one sentence naming the specific supernodes or features>"\},\\
~~"abstraction\_beyond\_tokens": \{"score": <integer 1-10>, "reason": "<one sentence naming the specific supernodes or features>"\},\\
~~"unexpected\_mechanism": \{"score": <integer 1-10>, "reason": "<one sentence naming the specific supernodes or features>"\},\\
~~"other\_interesting\_structure": \{"score": <integer 1-10>, "reason": "<one sentence naming the specific supernodes or features>"\}\\
\}
\end{promptbox}

\subsection{Validation Prompts}
\label{app:prompt_validation}

Each validation protocol uses a per-task prompt built from a template at scoring time. Runtime-substituted placeholders in the templates below: \texttt{\{group\_name\}} is the group description being probed; \texttt{\{description\}} is the feature description; \texttt{\{evidence\}} is a feature's raw evidence (input activations and output tokens); \texttt{\{n\_pos\}} is the number of correct items embedded among distractors (5 for text detection); \texttt{\{n\}} is the total number of candidates shown; the numbered list (\texttt{\{i. item\}}) is the shuffled candidates.

\begin{promptbox}[title=Feature Detection (1-in-10 feature ID): pick the feature that belongs to the group]
\begin{lstlisting}[style=promptstyle]
Group: "{group_name}"

Below are {n} feature descriptions from neurons inside a language model. Exactly 1 of these belongs to the group above.

Pick the single feature description that best matches the group.

1. {item_1}
2. {item_2}
...
{n}. {item_n}

Respond with the 1-based index of the best match.
\end{lstlisting}
\end{promptbox}

\begin{promptbox}[title=Text Detection (5-in-10 snippet match): pick the snippets that activated this group]
\begin{lstlisting}[style=promptstyle]
Group: "{group_name}"

Below are {n} text excerpts that strongly activated neurons inside a language model. The key activating tokens are highlighted with <<<>>>. Exactly {n_pos} of these come from neurons in this group.

Identify which {n_pos} text excerpts belong to this group.

1. {item_1}
2. {item_2}
...
{n}. {item_n}

Respond with the 1-based indices of the {n_pos} excerpts from this group.
\end{lstlisting}
\end{promptbox}

\begin{promptbox}[title=Description Detection (1-in-10 description ID): pick the description matching this feature's evidence]
\begin{lstlisting}[style=promptstyle]
Below is evidence about a single feature neuron in a language model (its input activations and output tokens).

{evidence}

Below are {n} candidate descriptions. Exactly 1 of these correctly describes this feature. Pick the single best match.

1. {item_1}
2. {item_2}
...
{n}. {item_n}

Respond with the 1-based index of the description that best matches the feature evidence above.
\end{lstlisting}
\end{promptbox}

\begin{promptbox}[title=Description Text Detection (5-in-10 snippet match): pick the snippets that activated this feature]
\begin{lstlisting}[style=promptstyle]
Feature description: "{description}"

Below are {n} text excerpts that strongly activated neurons inside a language model. The key activating tokens are highlighted with <<<>>>. Exactly {n_pos} of these come from the neuron described above.

Identify which {n_pos} text excerpts match this feature description.

1. {item_1}
2. {item_2}
...
{n}. {item_n}

Respond with the 1-based indices of the {n_pos} excerpts that activated this feature.
\end{lstlisting}
\end{promptbox}

\subsection{Feature Description Examples}
\label{tab:feature-descriptions}

Representative feature descriptions produced by the pipeline on a single prompt. Each row is one transcoder feature with its description LLM-generated from the top activating snippets and logits. The elaboration is from human analysis of the features themselves, looking at the top and bottom logits, as well as text snippets.

\begin{table}[hbt!]
  \centering
  \small
  \begin{tabular}{p{7cm}p{4.5cm}}
    \toprule
    Label & Elaboration \\
    \midrule
    Texas (state) — activates on mentions of Texas and Texas place names (cities, counties); tends to push output tokens like “Texas”, “TX”, “Texan” & All text snippets discuss Texas locations, with triggers predominantly being ``Texas.''\\
    capital (economics) — activates on the token "capital" in finance/production contexts; fires for loans, capital markets, raising capital, instruments of production. & The description is faithful. Triggers are predominantly the word ``capital'' with surrounding context discussing ``capital'' in economic and social contexts. \\
     City / city-list — activates on major city names and list contexts (commas, “and”); consistently promotes city tokens like Seattle, Atlanta, Chicago, London, “cities.” & The feature is polysemantic: triggered on various cities and city names like New York City, Mumbai, Chicago, London, and more. \\
     say capital city — fires on prepositions/articles (in, from, under, the, to) that precede locations; pushes capital/place names (Washington, Beijing, Canberra, "berra"). & The feature is often triggered on preposition tokens while the subsequent word is largely locations or capitals. \\
    say "governor" — activates on state names/“State”/state-official contexts; promotes the token governor/Governor and suppresses federal/national-related tokens & The top promoted tokens are all some form of ``governor'' or ``govern'' \\
    Dallas (city) — activates on mentions of Dallas and nearby local proper nouns (e.g., “Dallas Morning News”, local institutions); signals city/location context and local-entity mentions. & Most trigger tokens are ``Dallas,'' with addition weaker mentions of Texas and Texas counties. \\
    \bottomrule
  \end{tabular}
  \caption{Sample feature descriptions from a single attribution graph. ``The capital of the state containing Dallas is''}
\end{table}

\FloatBarrier

\section{Validation Protocol Details}
\label{app:validation_details}

\subsection{Feature-Level Protocols}
\label{app:d1d2}

The two feature-level protocols mirror the supernode-level pair but probe a single feature's description.

\begin{description}[leftmargin=*, style=nextline, itemsep=2pt, topsep=2pt]
  \item[Description Detection (Feature): specificity; 1-in-10 description ID.] For each feature, present its raw evidence (text snippet activations + trigger tokens) and ask the model to pick the correct description from 10 candidates (9 from other features). Random chance accuracy: 10\%.
  \item[Text Detection (Feature): sensitivity; 5-in-10 snippet match.] Given a feature's description, present 10 text snippets (5 from this feature's top activations, 5 from other features). The model picks the 5 that activated this feature. Random chance accuracy: 50\%.
\end{description}

These scores are per-feature and independent of grouping conditions; they apply identically across all Ours rows of Table~\ref{tab:autointerp_vs_human}, so we report only Ours (full) below.

\begin{table}[hbt!]
  \centering
  \small
  \begin{tabular}{lcc}
    \toprule
    Condition & \shortstack{Desc.\ Det.\\(Feature) ($\uparrow$)} & \shortstack{Text Det.\\(Feature) ($\uparrow$)} \\
    \midrule
    Ours (full)  & 79.9\% & 85.4\% \\
    Chance       & 10\%   & 50\%   \\
    \bottomrule
  \end{tabular}

  \caption{\textbf{Per-feature description scores.} Description Detection (Feature) and Text Detection (Feature) probe the feature description rather than the supernode description. Scores are constant across the top-$k$ and refine/no-refine variants of Ours, so we report only Ours (full).}
  \label{tab:d1d2_scores}
\end{table}

\subsection{Hop Matching}
\label{app:hop-matching}

For each graph in the multi-hop experiments, we record (i) whether the model is top-1 correct and (ii) whether the intermediate hop supernode is present. Hop presence is determined by lowercase string and stem matching against supernode labels: e.g., an ``Islam'' answer matches supernodes named ``say Islam'' or ``Islam location.'' The same matching logic is reused for intervention and amplification. False positives (e.g.\ ``University'' matching with ``University of Budapest'') are manually confirmed and removed from all results; false negatives (e.g.\ ``Portuguese'' vs.\ ``Portugal'') are excluded from intervention and amplification numbers, but are recovered manually for Table~\ref{tab:capitals-hop-outcomes}. All results are free of hop matching issues.

\subsection{Score Aggregation}
\label{validation_aggregate}

Each validation question (Feature Detection or Text Detection) is run with 5 random seeds to stabilize the estimate. For each prompt, we average across its supernodes; we then average across the 15 prompts. This weights each prompt equally regardless of how many supernodes it has.

\subsection{Validation Consistency Across Models}
\label{validation_models}
\begin{table}[hbt!]
  \centering
  \small
  \begin{tabular}{lcccccc}
    \toprule
    & \multicolumn{2}{c}{GPT-5 mini} & \multicolumn{2}{c}{Gemini 2.5 Flash} & \multicolumn{2}{c}{Claude Haiku 4.5} \\
    \cmidrule(lr){2-3}\cmidrule(lr){4-5}\cmidrule(lr){6-7}
    Condition & \shortstack{Feat.\\Det.} & \shortstack{Text\\Det.} & \shortstack{Feat.\\Det.} & \shortstack{Text\\Det.} & \shortstack{Feat.\\Det.} & \shortstack{Text\\Det.} \\
    \midrule
    Random baseline     & 22.1\% & 54.7\% & 22.7\% & 54.2\% & 19.7\% & 50.3\% \\
    Human-annotated     & 66.6\% & 81.8\% & 65.7\% & 81.9\% & 58.3\% & 72.0\% \\
    Ours (no refine) & 92.3\% & 82.1\% & 91.8\% & 80.1\% & 83.9\% & 70.5\% \\
    Ours (full)         & 93.4\% & 83.2\% & 93.3\% & 81.3\% & 84.9\% & 70.9\% \\
    \midrule
    Chance              & \multicolumn{2}{c}{10\% / 50\%} & \multicolumn{2}{c}{10\% / 50\%} & \multicolumn{2}{c}{10\% / 50\%} \\
    \bottomrule
  \end{tabular}
  
  \caption{Feature Detection and Text Detection validation scores across three different judge models on the 15 Neuronpedia \citet{circuit-tracer} graphs. Results are consistent across GPT-5 mini and Gemini-2.5-flash; Claude Haiku shows slightly lower absolute scores but preserves roughly the same ordering. Judge bias is not a significant confounder of validation results.}
  \label{tab:cross-model-validation}
\end{table}

\FloatBarrier
\subsection{Validation Consistency Across Group Sizes}
\label{validation sizes}

\begin{table}[hbt!]
  \centering
  \small
  \begin{tabular}{lccccc}
    \toprule
    Condition & \shortstack{Feature\\Detection} ($\uparrow$) & \shortstack{Text\\Detection} ($\uparrow$) & \shortstack{Features\\grouped} & \shortstack{\#\\Groups} & \shortstack{Features/\\supernode} \\
    \midrule
    Human-annotated              & 66.6\% & 81.8\% & 33.7   & 6.6  & 4.91  \\
    Ours (top 50, no refine)  & 82.1\% & 81.7\% & 30.3   & 8.7  & 3.50  \\
    Ours (top 50)                & 80.8\% & 81.7\% & 26.9   & 7.1  & 3.98  \\
    Ours (top 100, no refine) & 82.4\% & 80.6\% & 54.9   & 8.7  & 6.43  \\
    Ours (top 100)               & 83.1\% & 82.1\% & 48.5   & 7.1  & 7.08  \\
    Ours (top 150, no refine) & 81.7\% & 81.7\% & 77.1   & 9.3  & 8.60  \\
    Ours (top 150)               & 82.8\% & 84.2\% & 68.1   & 7.5  & 9.43  \\
    Ours (top 200, no refine) & 83.2\% & 81.1\% & 97.6   & 9.7  & 10.51 \\
    Ours (top 200)               & 81.9\% & 81.7\% & 88.5   & 8.1  & 11.62 \\
    Ours (no refine)          & 92.3\% & 82.1\% & 250.1  & 11.1 & 24.30 \\
    Ours (full)                  & 93.4\% & 83.2\% & 207.5  & 8.3  & 26.42 \\
    \midrule
    Random baseline              & 22.1\% & 54.7\% & --     & --   & --    \\
    Chance                       & 10\%   & 50\%   & --     & --   & --    \\
    \bottomrule
  \end{tabular}
  
  \caption{\textbf{Autointerp scores: full table including pre-refine variants.} Same setup as Table~\ref{tab:autointerp_vs_human}. The refine pass contributes a small, consistent improvement on Feature Detection across top-$k$ settings.}
  \label{tab:autointerp_full}
\end{table}

\begin{table}[hbt!]
  \centering
  \small
  \begin{tabular}{lcccccc}
    \toprule
    & \multicolumn{3}{c}{Feature Detection ($\uparrow$)} & \multicolumn{3}{c}{Text Detection ($\uparrow$)} \\
    \cmidrule(lr){2-4} \cmidrule(lr){5-7}
    Condition & $k\geq 2$ & $k\geq 3$ & $k\geq 4$ & $k\geq 2$ & $k\geq 3$ & $k\geq 4$ \\
    \midrule
    Random baseline          & 22.1\% & 21.9\% & 21.4\% & 54.7\% & 54.8\% & 54.9\% \\
    Human-annotated          & 66.6\% & 66.7\% & 69.2\% & 81.8\% & 82.3\% & 82.9\% \\
    Ours (no refine)      & 92.3\% & 91.6\% & 91.0\% & 82.1\% & 80.3\% & 80.6\% \\
    Ours (full)              & 93.4\% & 92.2\% & 92.9\% & 83.2\% & 81.8\% & 81.8\% \\
    \bottomrule
  \end{tabular}
  
  \caption{Stability of Feature Detection and Text Detection scores across minimum supernode-size thresholds $k$ on the 15 Neuronpedia graphs from \citet{circuit-tracer}. The headline numbers in Table~\ref{tab:autointerp_vs_human} ($k\geq 2$) are not an artifact of the threshold choice. Results change minimally across supernode sizes.}
  \label{tab:size_stability}
\end{table}

\FloatBarrier
\section{Dataset Construction}
\label{app:datasets}

\subsection{Replication of \citet{circuit-tracer} Reference Graphs}
\label{neuronpedia prompts}
We use the 15 human-annotated reference graphs from the \texttt{circuit-tracer} Gemma demo\footnote{\url{https://github.com/decoderesearch/circuit-tracer/blob/main/demos/gemma_demo.ipynb}}~\citep{circuit-tracer}, hosted on Neuronpedia \citep{neuronpedia}.

\begin{longtable}{p{10cm}c}
  \toprule
  Prompt & Answer \\
  \midrule
  \endfirsthead
  \multicolumn{2}{c}{\textit{(Neuronpedia, continued)}} \\
  \toprule
  Prompt & Answer \\
  \midrule
  \endhead
  \bottomrule
  \endfoot
    The International Advanced Security Group (IAS & G \\
    Mexico:peso :: Europe: & euro \\
    The guitarist knew the song & . / is \\
    The keys on the cabinet & are \\
    Fact: Michael Jordan plays the sport of & basketball \\
    La saison après le printemps s'apelle l' & été \\
    La estación después de la primavera se llama el & verano \\
    The girl that the teacher sees & is \\
    The girls that the teacher sees & are \\
    Fait: Michael Jordan joue au & basket \\
    Hecho: Michael Jordan juega al & baloncesto \\
    2 + 1 = & 3 \\
    3 + 5 = & 8 \\
    Mexico:Spanish :: US: & English \\
    Mexico:peso :: US: & dollar \\
  \caption{All 15 \citet{circuit-tracer} replication prompts.}\label{tab:all-neuronpedia} \\
\end{longtable}

\subsection{Capitals}
100 prompts following the template ``The capital of the \{state/country\} containing \{city\} is''.
The intermediate concept is a US state (50 prompts) or a non-US country (50 prompts).

\begin{table}[hbt!]
  \centering
  \small
  \begin{tabular}{p{6.5cm}cc}
    \toprule
    Prompt & Intermediate & Answer \\
    \midrule
    The capital of the state containing Birmingham is & Alabama & Montgomery \\
    The capital of the state containing Anchorage is & Alaska & Juneau \\
    The capital of the state containing the Grand Canyon is & Arizona & Phoenix \\
    The capital of the state containing Fort Smith is & Arkansas & Little Rock \\
    The capital of the state containing Los Angeles is & California & Sacramento \\
    The capital of the country containing Toronto is & Canada & Ottawa \\
    The capital of the country containing Cancun is & Mexico & Mexico City \\
    The capital of the country containing Manchester is & United Kingdom & London \\
    The capital of the country containing Munich is & Germany & Berlin \\
    The capital of the country containing Mumbai is & India & New Delhi \\
    \multicolumn{3}{c}{\ldots\ (90 more)} \\
    \bottomrule
  \end{tabular}
  \caption{Sample of 10 Capitals prompts (5 state-based, 5 country-based). Full 100-prompt list in our code repository.\textsuperscript{\ref{fn:code}}}
  \label{tab:all-capitals}
\end{table}

\subsection{Wikipedia}
\label{app:wiki-construction}
\paragraph{Construction.}
1000 mid-sentence completions from random Wikipedia articles (8--15 word prefixes): 500 one-sentence prompts (a single mid-sentence prefix) and 500 context prompts (a complete preceding sentence prepended to the prefix). The number of prompts is a configurable target; the candidate pool we fetch and score scales with that target (the selection yield is only a few percent), so it is not a fixed count. We keep positions where: (1) Gemma-2-2B's top-1 token matches the true next word; (2) the top-1 probability is at most $0.8$; (3) the target token is not a stop-word, article, punctuation, or short/non-ASCII subword fragment; and (4) the prompt does not end on a word that syntactically forces a function word next (e.g.\ ``of'', ``the''). We cap each article at five prompts for diversity. No intermediate concept applies; the target column is the true Wikipedia next token.

Articles are seeded from topic searches across six broad categories,
then supplemented with random Wikipedia pages until the candidate pool reaches the
target size.

\begin{table}[hbt!]
  \centering
  \small
  \begin{tabular}{lp{8.5cm}}
    \toprule
    Category & Example seed topics \\
    \midrule
    Science \& Technology   & History of science, Neuroscience, Astronomy, Mathematics, Computer science, Artificial intelligence, Space exploration, Physics, Chemistry, Biology \\
    History \& Civilization & World War II, Ancient Rome, French Revolution, Industrial Revolution, Cold War, Ottoman Empire, British Empire, Chinese dynasties \\
    Arts \& Culture         & Renaissance art, Jazz music, Classical music, Film history, Literature, Architecture, Impressionism, Opera \\
    Geography \& Nature     & Amazon rainforest, Solar System, Island nations, Mountain ranges, Desert ecosystems, Major rivers \\
    Social Sciences         & Linguistics, Economics, Sociology, Anthropology, Psychology, Political philosophy \\
    Sports \& Other         & Olympic Games, History of chess, History of sport \\
    \bottomrule
  \end{tabular}
  \caption{Seed topic categories used for Wikipedia article sampling.}
  \label{tab:wiki-categories}
\end{table}

\begin{table}[hbt!]
  \centering
  \small
  \begin{tabular}{p{6.5cm}cc}
    \toprule
    Prompt prefix & Target token & Conf. \\
    \midrule
    Booth in 1901, Scrooge, or, Marley's Ghost is the earliest film & adaptation & 0.399 \\
    Some of the best candidates for future deep & space & 0.435 \\
    John Westergaard's and Henrietta Resler's Class in a capitalist & society & 0.352 \\
    Researchers in educational neuroscience investigate the neural mechanisms of reading, numerical & cognition & 0.270 \\
    Australia is not included as it is considered a continental & country & 0.395 \\
    The Babylonians discovered that lunar eclipses recurred in the saros & cycle & 0.389 \\
    Beyond AGI, artificial superintelligence (ASI) would outperform the best human abilities across every & domain & 0.417 \\
    The myth continues that after Heracles completed his twelve labours, he built the Olympic & stadium & 0.379 \\
    Models in theoretical neuroscience are aimed at capturing the essential & features & 0.303 \\
    The Cold War also saw a nuclear arms race between the two & superpowers & 0.501 \\
    \multicolumn{3}{c}{\ldots\ (490 more)} \\
    \bottomrule
  \end{tabular}
  \caption{Sample of 10 one-sentence Wikipedia prompts. Full 500-prompt list in our code repository.\textsuperscript{\ref{fn:code}}}
  \label{tab:all-wikipedia}
\end{table}

\begin{table}[hbt!]
  \centering
  \small
  \begin{tabular}{p{6.5cm}cc}
    \toprule
    Prompt prefix & Target token & Conf. \\
    \midrule
    In his keynote address, printed in the first volume of its new publication, The Anthropological Review, Hunt stressed the work of Waitz, adopting his definitions as a standard. Among the first associates were the young Edward Burnett Tylor, inventor of cultural & anthropology & 0.410 \\
    In 2016, it was recognized that the first possible evidence of an exoplanet had been noted in 1917. As of 23 April 2026, there are 6,273 & confirmed & 0.436 \\
    Homeric Greek had significant differences in grammar and pronunciation from Classical Attic and other Classical-era dialects. The origins, early form and development of the Hellenic & language & 0.400 \\
    All depended upon farmers producing an agricultural surplus to support the centralized government, political leaders, religious leaders, and public works of the urban centers of the early civilizations. The earliest signs of a process leading to sedentary culture can be & seen & 0.399 \\
    Climate change may occur over long and short timescales due to various factors. Recent warming is discussed in terms of global & warming & 0.398 \\
    The Romance family itself is part of the larger Indo-European family, which includes many other languages native to Europe and South Asia, all believed to have descended from a common ancestor known as Proto-Indo-European. A language family is usually said to contain at least two & languages & 0.397 \\
    Its native distribution is in southern and western Europe and North Africa. It occurs as a scarce plant in south-west & england & 0.405 \\
    Therefore, they are able to inhabit many types of geography and topology. Research suggests that these spiders prefer flatter and more & open & 0.406 \\
    Climate change mitigation actions include conserving energy and replacing fossil fuels with clean energy sources. Secondary mitigation strategies include changes to land use and removing & carbon & 0.394 \\

    \multicolumn{3}{c}{\ldots\ (490 more)} \\
    \bottomrule
  \end{tabular}
  \caption{Sample of 10 context Wikipedia prompts, each a complete sentence followed by an incomplete second sentence. Full 500-prompt list in our code repository.\textsuperscript{\ref{fn:code}}}
  \label{tab:all-wikipedia-context}
\end{table}

\FloatBarrier
\section{Probing the Grouping Pipeline on Arithmetic}
\label{app:math}

One concern is that supernodes may often be too vague to be informative. We test this on arithmetic. Prior work~\citep{lindsey2025biology} has shown that language models add numbers via two parallel pathways: one computes exact single-digit addition, the other estimates the rough magnitude of the result. We apply our pipeline to $\sim$200 two- and three-digit addition prompts.

\begin{figure}[hbt!]
    \centering
    \includegraphics[width=0.53\linewidth]{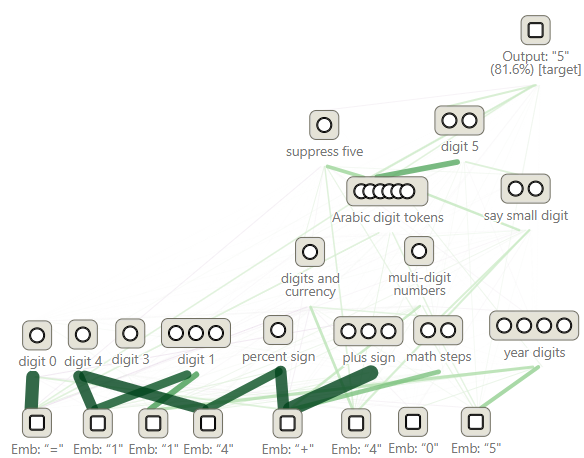}
    \hfill
    \includegraphics[width=0.45\linewidth]{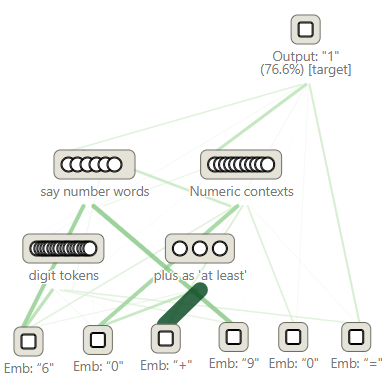}

    \caption{\textbf{Two example arithmetic graphs} (left: \texttt{114+405=}$\to$``5''; right: \texttt{60+90=}$\to$``1''). Supernodes are either surface-token groups (\textsc{digit 1/4}, \textsc{plus sign}, \textsc{digit tokens}) or generic numeric-context groups (\textsc{year digits}, \textsc{digits and currency}, \textsc{plus as `at least'}), with no carrying, magnitude, or lookup step.}
    \label{fig:math_figures}
\end{figure}

We show two annotated attribution graphs in \Cref{fig:math_figures}. Qualitatively, the resulting supernodes are either at too low a level, labeling specific digits in the input, or too high a level, with vague labels like ``numeric context.'' The labels are too coarse to reveal the structure that prior work has shown exists at the feature level.

We quantify this with an LLM judge that categorizes each supernode name across all graphs (\Cref{tab:math-taxonomy}). Only 12 of 1885 supernodes (1\%) are classified as naming an arithmetic mechanism.

\begin{table}[hbt!]
  \centering
  \small
  \begin{tabular}{lrrr}
    \toprule
    Bucket & Instances & Share & Unique names \\
    \midrule
    surface-token & 1228 & 65\% & 368 \\
    say-number    & 542  & 29\% & 191 \\
    mechanism     & 12   & 1\%  & 9   \\
    other         & 103  & 5\%  & 81  \\
    \bottomrule
  \end{tabular}
  \caption{\textbf{Supernode-name taxonomy on the math graphs.} Group names overwhelmingly identify surface tokens or generic number output; almost none name an arithmetic mechanism.}
  \label{tab:math-taxonomy}
\end{table}

Despite these supernodes being too coarse to reveal this known arithmetic structure, they score well on the autointerp metrics from \Cref{sec:group_validation} (\Cref{tab:math-validation}). This suggests that autointerp scores measure whether a grouping is internally coherent, not whether it reveals meaningful structure. A natural direction for future work is to tune automatic annotation pipelines to recover this known structure. We hypothesize that our general approach of describing features and then grouping descriptions could still work, with further refinement of the description and grouping steps.

\begin{table}[hbt!]
  \centering
  \small
  \begin{tabular}{lcc}
    \toprule
    Metric & Mean & Chance \\
    \midrule
    Feature Detection (Supernode)    & 87\% & 10\% \\
    Text Detection (Supernode)       & 70\% & 50\% \\
    Description Detection (Feature)  & 73\% & 10\% \\
    Text Detection (Feature)         & 81\% & 50\% \\
    \bottomrule
  \end{tabular}
  \caption{\textbf{Autointerp validation on 20 math graphs.} The groups validate well (they are internally coherent) despite carrying weak mechanistic content for the arithmetic task.}
  \label{tab:math-validation}
\end{table}

\FloatBarrier
\section{Graph Gallery}
\label{app:gallery}

\subsection{\citet{circuit-tracer} Replication: Side-by-Side}
\label{Neuronpedia sidebyside}
8 representative prompts from the 15-prompt \citet{circuit-tracer} replication set (Appendix~\ref{neuronpedia prompts}).

\begin{figure}[hbt!]
    \centering
    \includegraphics[width=0.48\linewidth]{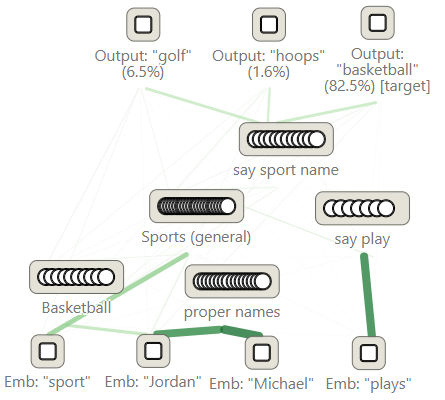}
    \hfill
    \includegraphics[width=0.48\linewidth]{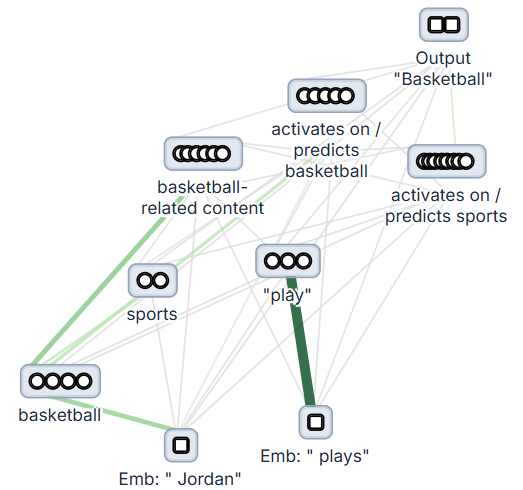}
    
    \caption{Left: our LLM-grouped supernodes (top 100). Right: \citet{circuit-tracer} human-annotated supernodes. Both pipelines surface an explicit Basketball intermediate.}
    \label{fig:mega_figure1}
\end{figure}

\begin{figure}[hbt!]
    \centering
    \includegraphics[width=0.48\linewidth]{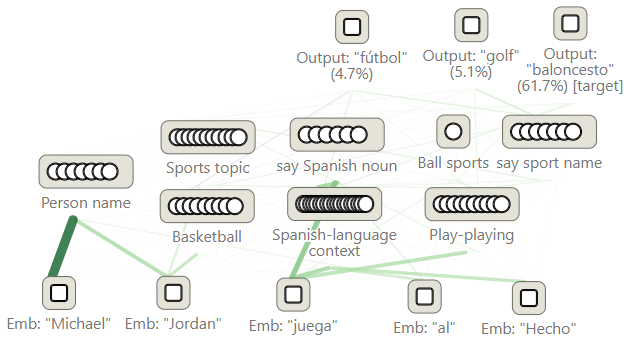}
    \hfill
    \includegraphics[width=0.48\linewidth]{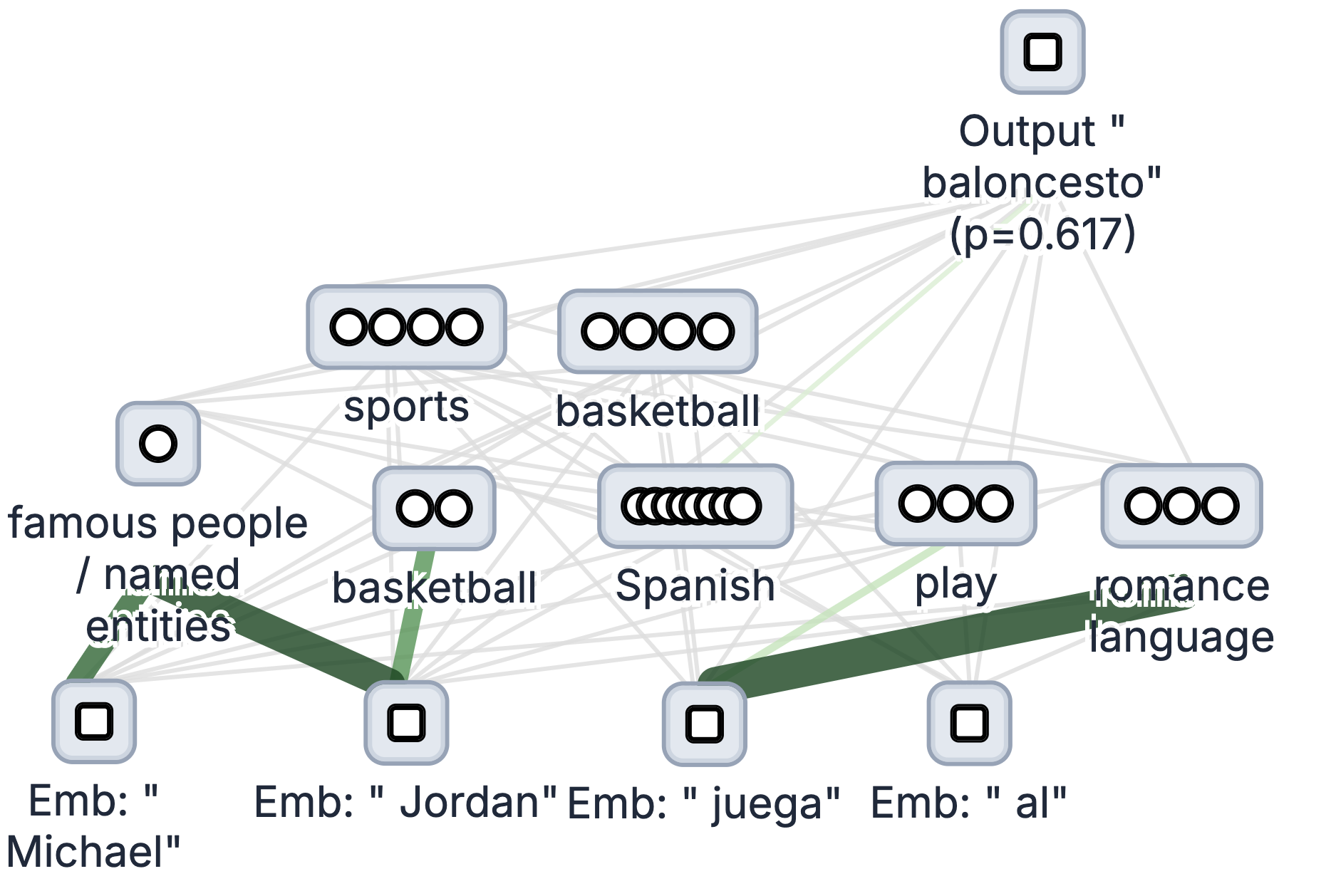}
    
    \caption{Left: ours (top 100). Right: \citet{circuit-tracer} human-annotated supernodes. The Michael Jordan → basketball hop is recovered in both, and on top of it our grouping isolates a Spanish-language context supernode that captures the cross-lingual transfer separately from the sport concept.}
    \label{fig:mega_figure2}
\end{figure}

\begin{figure}[hbt!]
    \centering
    \includegraphics[width=0.48\linewidth]{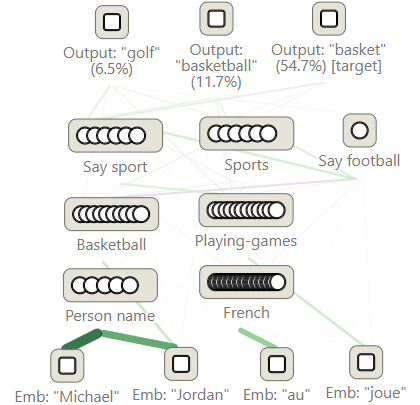}
    \hfill
    \includegraphics[width=0.48\linewidth]{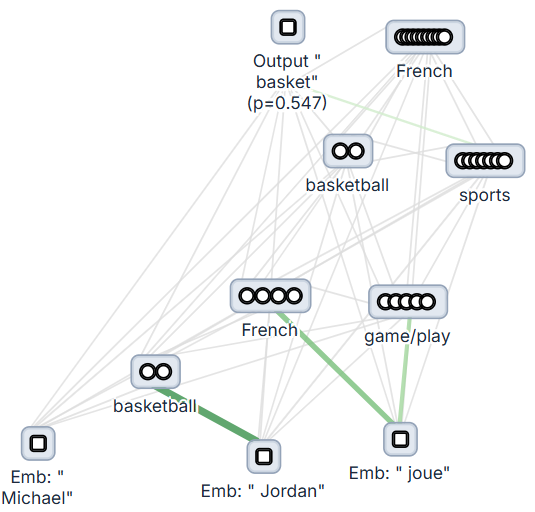}

    \caption{Left: ours (top 100). Right: \citet{circuit-tracer} human-annotated supernodes. The same MJ→basketball circuit appears under a French output token. Ours additionally surfaces a Say football group competing with the target output.}
    \label{fig:mega_figure3}
\end{figure}

\begin{figure}[hbt!]
    \centering
    \includegraphics[width=0.48\linewidth]{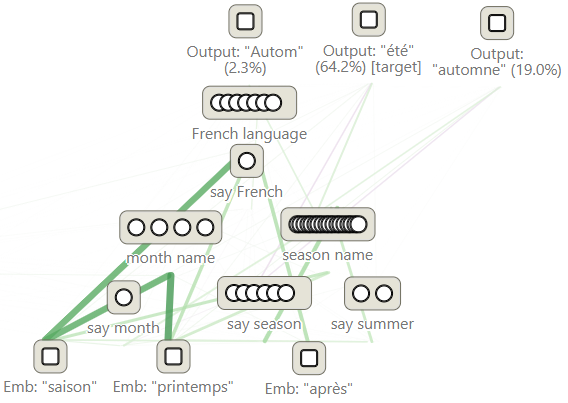}
    \hfill  
    \includegraphics[width=0.48\linewidth]{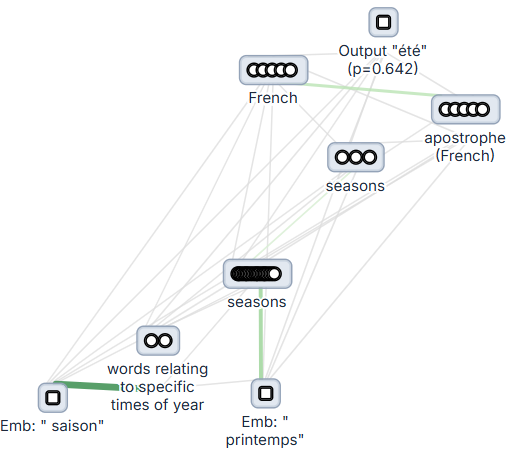}

    \caption{Left: ours (top 100). Right: \citet{circuit-tracer} human-annotated supernodes. Our grouping surfaces an explicit say summer supernode that names the intermediate concept; the human grouping uses a more generic seasons / words relating to specific times of year cluster that does not call out the target season.}
    \label{fig:mega_figure4}
\end{figure}

\begin{figure}[hbt!]
    \centering
    \includegraphics[width=0.48\linewidth]{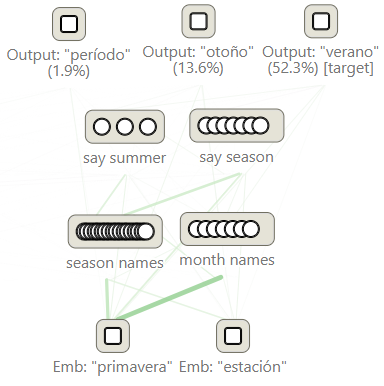}
    \hfill
    \includegraphics[width=0.48\linewidth]{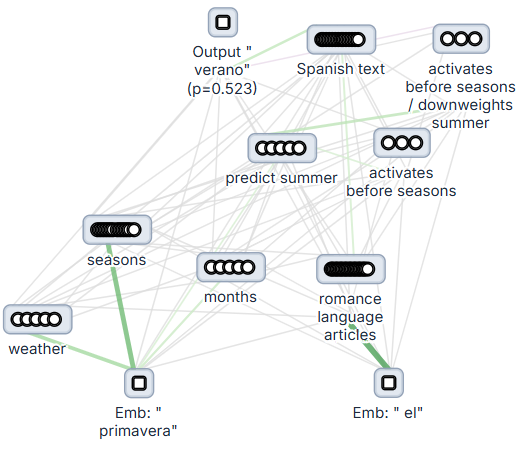}

    \caption{Left: ours (top 100). Right: \citet{circuit-tracer} human-annotated supernodes. Both find a predict/say summer intermediate.}
    \label{fig:mega_figure5}
\end{figure}

\begin{figure}[hbt!]
    \centering
    \includegraphics[width=0.48\linewidth]{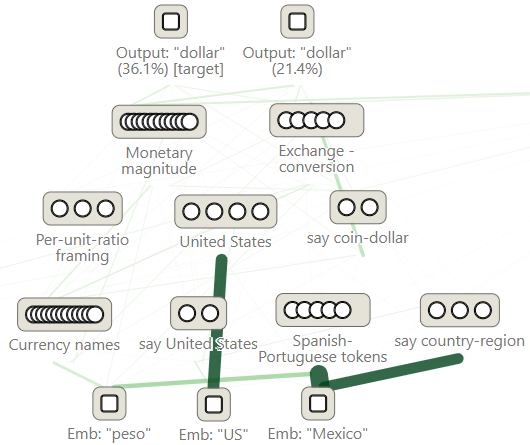}
    \hfill
    \includegraphics[width=0.48\linewidth]{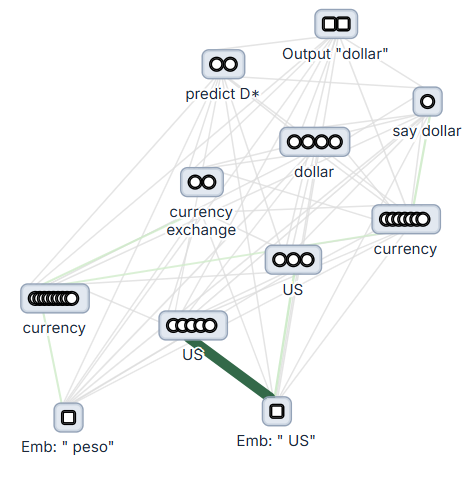}

    \caption{Left: ours (top 100). Right: \citet{circuit-tracer} human-annotated supernodes. The two-step analogy is recovered on both sides: a currency supernode encodes the relation, and a US supernode encodes the country mapping that selects "dollar" as its currency.}
    \label{fig:mega_figure6}
\end{figure}

\begin{figure}[hbt!]
    \centering   
    \includegraphics[width=0.48\linewidth]{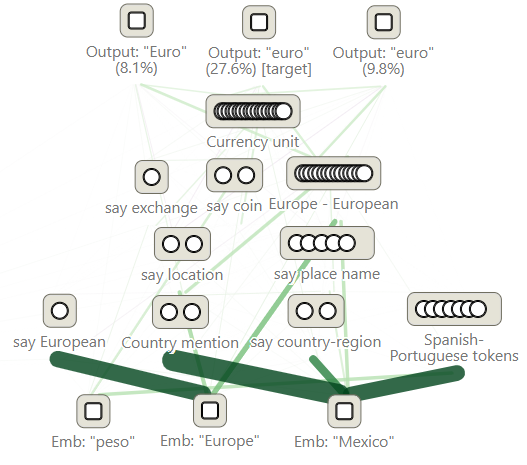}
    \hfill
    \includegraphics[width=0.48\linewidth]{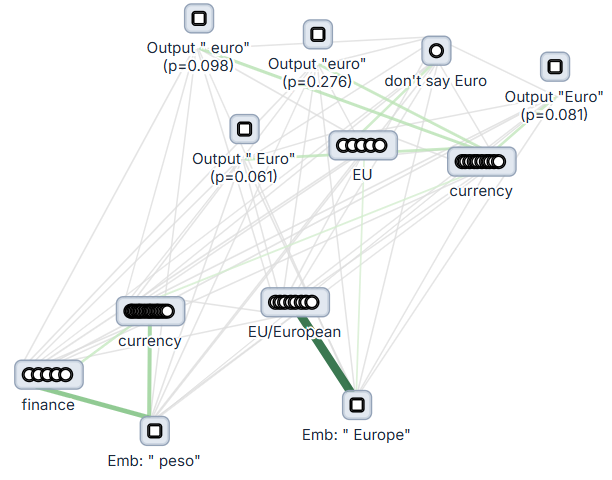}

    \caption{Left: ours (top 100). Right: \citet{circuit-tracer} human-annotated supernodes. Currency-analogy structure with the country term swapped to Europe / EU.}
    \label{fig:mega_figure7}
\end{figure}

\begin{figure}[hbt!]
    \centering 
    \includegraphics[width=0.48\linewidth]{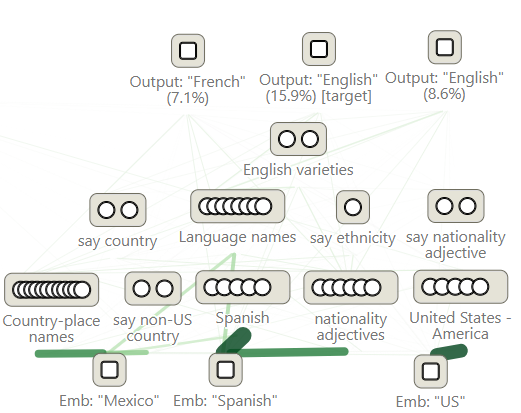}
    \hfill
    \includegraphics[width=0.48\linewidth]{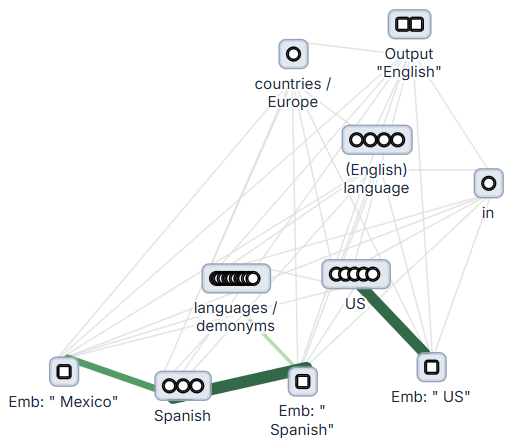}

    \caption{Left: ours (top 100). Right: \citet{circuit-tracer} human-annotated supernodes. The same analogy template applied to language rather than currency. Our grouping recovers a fine-grained set of country, nationality adjective, and language name supernodes, while \citet{circuit-tracer} consolidates these into broader clusters.}
    \label{fig:mega_figure8}
\end{figure}

\FloatBarrier
\subsection{Capitals: Canonical 2-Hop Examples}

Selected examples from the 100-prompt Capitals dataset annotated by our pipeline.

\begin{figure}[hbt!]
  \centering
  \begin{subfigure}{0.48\linewidth}
    \includegraphics[width=\linewidth]{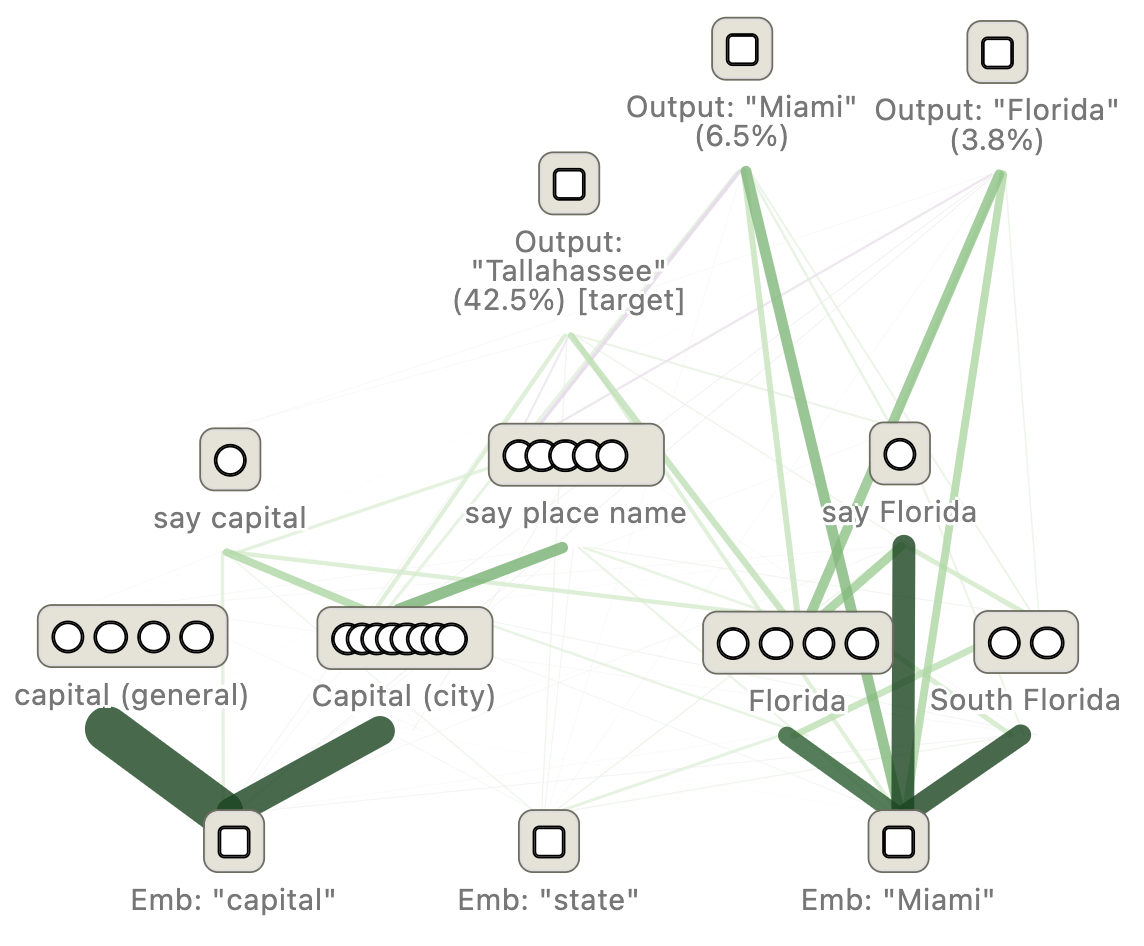}
    \caption{Miami $\to$ Florida $\to$ \emph{Tallahassee}.}
    \label{fig:gallery-cap1}
  \end{subfigure}\hfill
  \begin{subfigure}{0.48\linewidth}
    \includegraphics[width=\linewidth]{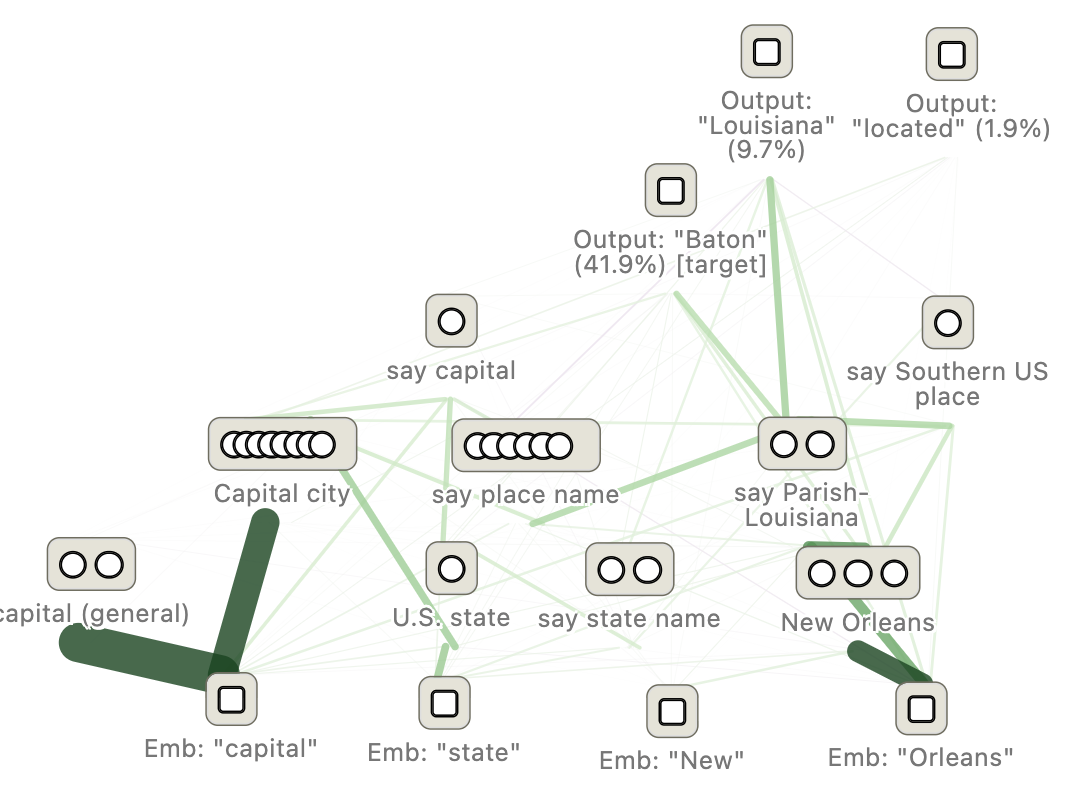}
    \caption{New Orleans $\to$ Louisiana $\to$ \emph{Baton Rouge}.}
    \label{fig:gallery-cap2}
  \end{subfigure}
  \vspace{4pt}
  \begin{subfigure}{0.48\linewidth}
    \includegraphics[width=\linewidth]{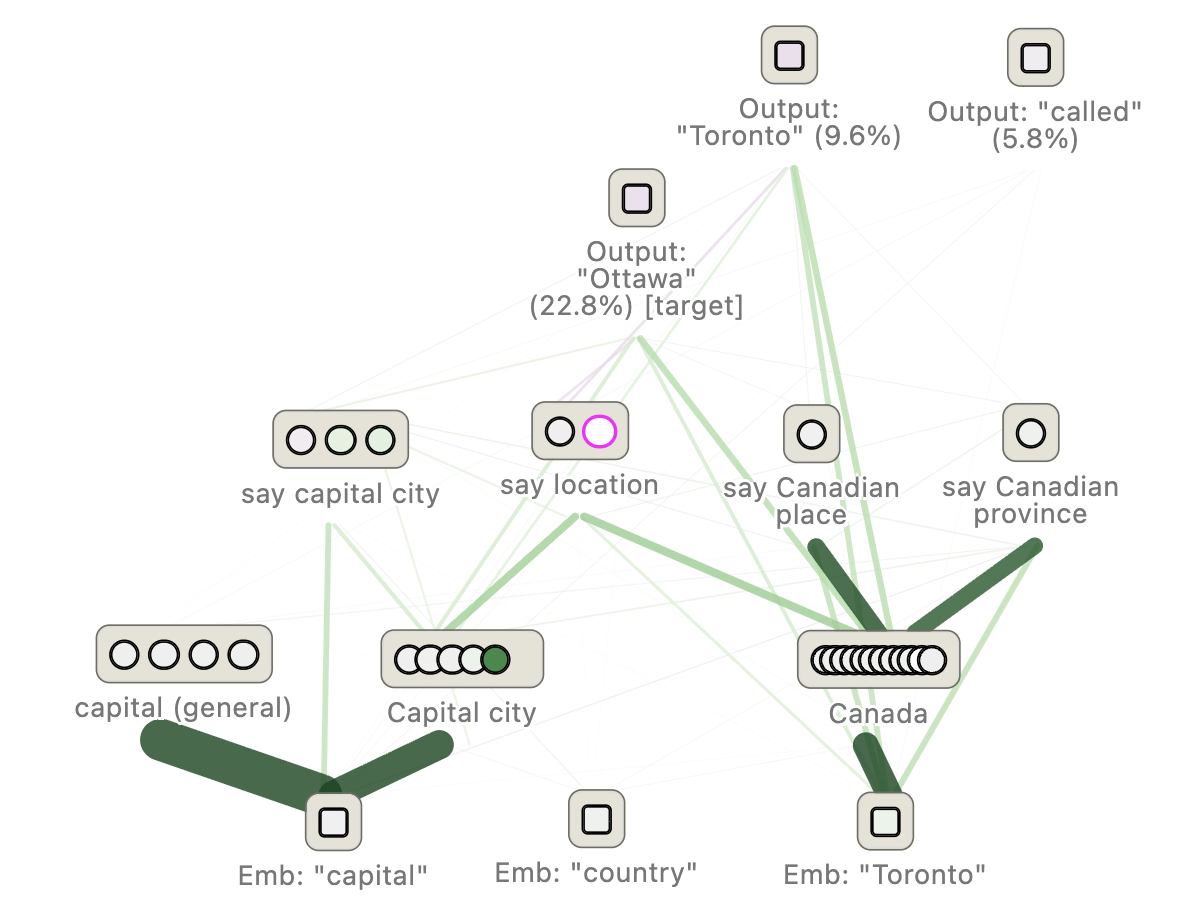}
    \caption{Toronto $\to$ Canada $\to$ \emph{Ottawa}.}
    \label{fig:gallery-cap3}
  \end{subfigure}\hfill
  \begin{subfigure}{0.48\linewidth}
    \includegraphics[width=\linewidth]{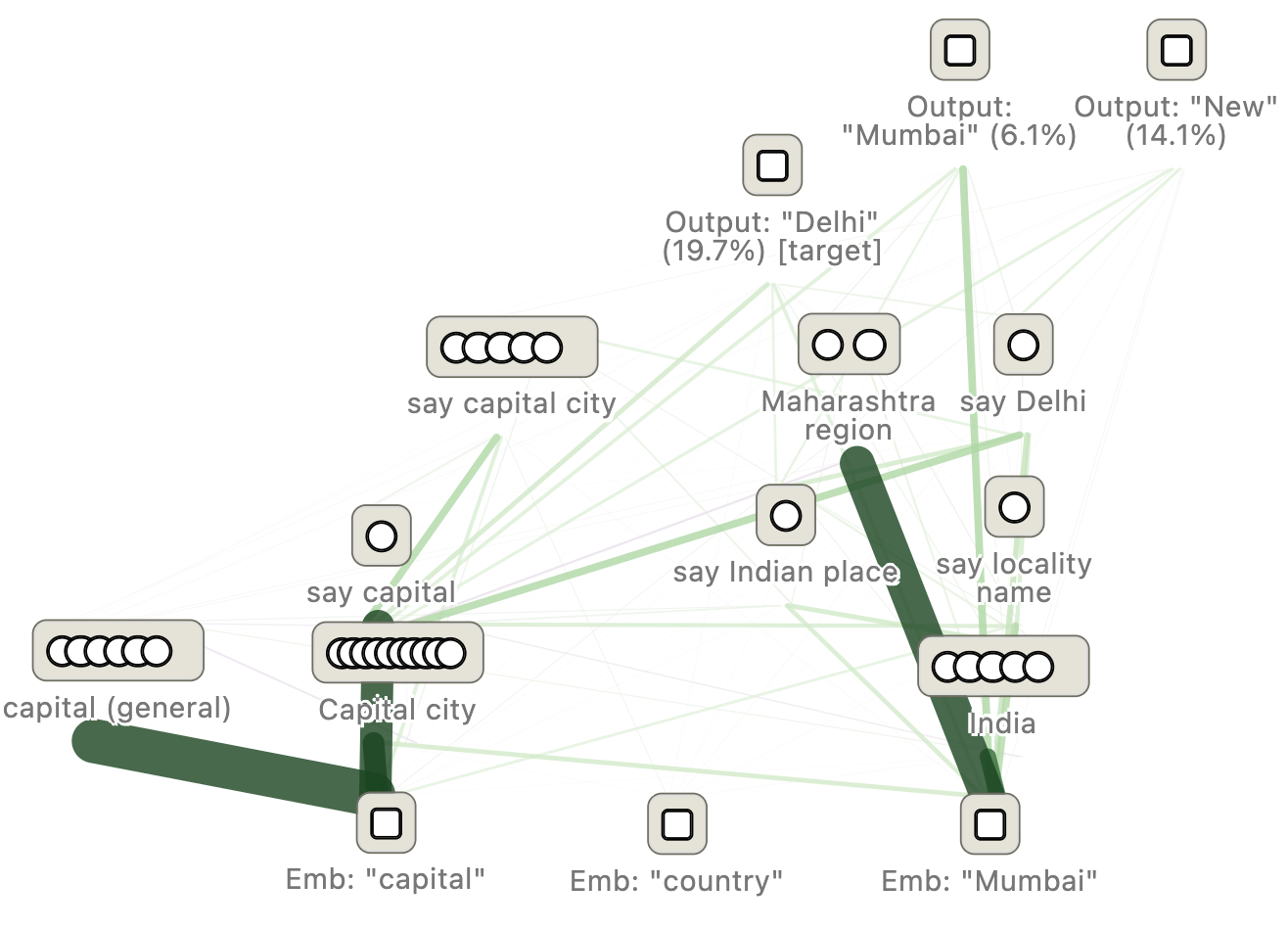}
    \caption{Mumbai $\to$ India $\to$ \emph{New Delhi}.}
    \label{fig:gallery-cap5}
  \end{subfigure}
  \caption{\textbf{Capitals 2-hop graphs.} Each prompt is of the form ``The capital of the \{state/country\} containing \{city\} is''. The intermediate state or country is the expected middle-hop supernode.}
  \label{fig:gallery-capitals}
\end{figure}

\FloatBarrier
\section{Automating MLP Neurons Versus SLT Features}
\label{app:mlp}

Our main results decompose Gemma-2-2B with Gemma Scope single-layer transcoder features. To test the generalization of our automation pipeline, we rerun it on the model's raw MLP neurons and compare the two decompositions on the 100 Capitals prompts.

\subsection{Setup}
\label{app:mlp-setup}

We build our attribution graphs over the model's individual MLP neurons rather than transcoder features using the recent ADAG codebase~\citep{arora2026adagautomaticallydescribingattribution}. We then apply the identical description and grouping pipeline (\Cref{sec:pipeline-desc,sec:pipeline-group}), with the same GPT-5-mini. 

\paragraph{Ensuring Fair Comparison.} MLP neurons are not hosted on Neuronpedia, so we collect activating-text exemplars ourselves, matching the corpus, token budget, and context length used to build Gemma Scope's dashboards. Additionally, with no hosted logits, we compute each neuron's most promoted and suppressed tokens using the same method as the transcoder logits. However, because an MLP neuron's activation is signed (unlike non-negative transcoder features), a neuron's positive- and negative-firing behavior can differ; we describe each separately and label every graph node with the description matching the sign of its activation. Lastly, we keep the top 150 neurons per graph after pruning; similar to transcoder graphs, we prune based on attribution magnitude (\Cref{sec:methods}).

\subsection{How Much Gets Grouped}
\label{app:mlp-grouping}

\begin{table}[hbt!]
  \centering
  \small
  \begin{tabular}{lcccc}
    \toprule
    Method & Grouped & Ungrouped & Total & Fraction grouped \\
    \midrule
    MLP neurons & 32.1 & 116.0 & 148.1 & 21.6\% \\
    Transcoder  & 26.8 & 17.8  & 44.6  & 60.2\% \\
    \bottomrule
  \end{tabular}
  \caption{\textbf{Grouped vs.\ ungrouped features per Capitals graph.} Counts are means per graph over the 100 Capitals graphs; the fraction grouped is pooled across all graphs. The pipeline places far fewer MLP neurons into supernodes (21.6\%) than transcoder features (60.2\%).}
  \label{tab:mlp-grouping}
\end{table}

Across the 100 Capitals graphs, the pipeline groups a much smaller fraction of MLP neurons into supernodes than transcoder features (21.6\% vs.\ 60.2\%; \Cref{tab:mlp-grouping}), suggesting many MLP neurons do not have a clean concept. Therefore, we use 150 neurons per graph, so that the number of MLP neurons actually placed into supernodes is comparable to what we see for transcoders: 32.1 per graph for MLP neurons versus 26.8 for SLT features.

\begin{figure*}[hbt!]
  \centering
  \includegraphics[width=0.48\textwidth]{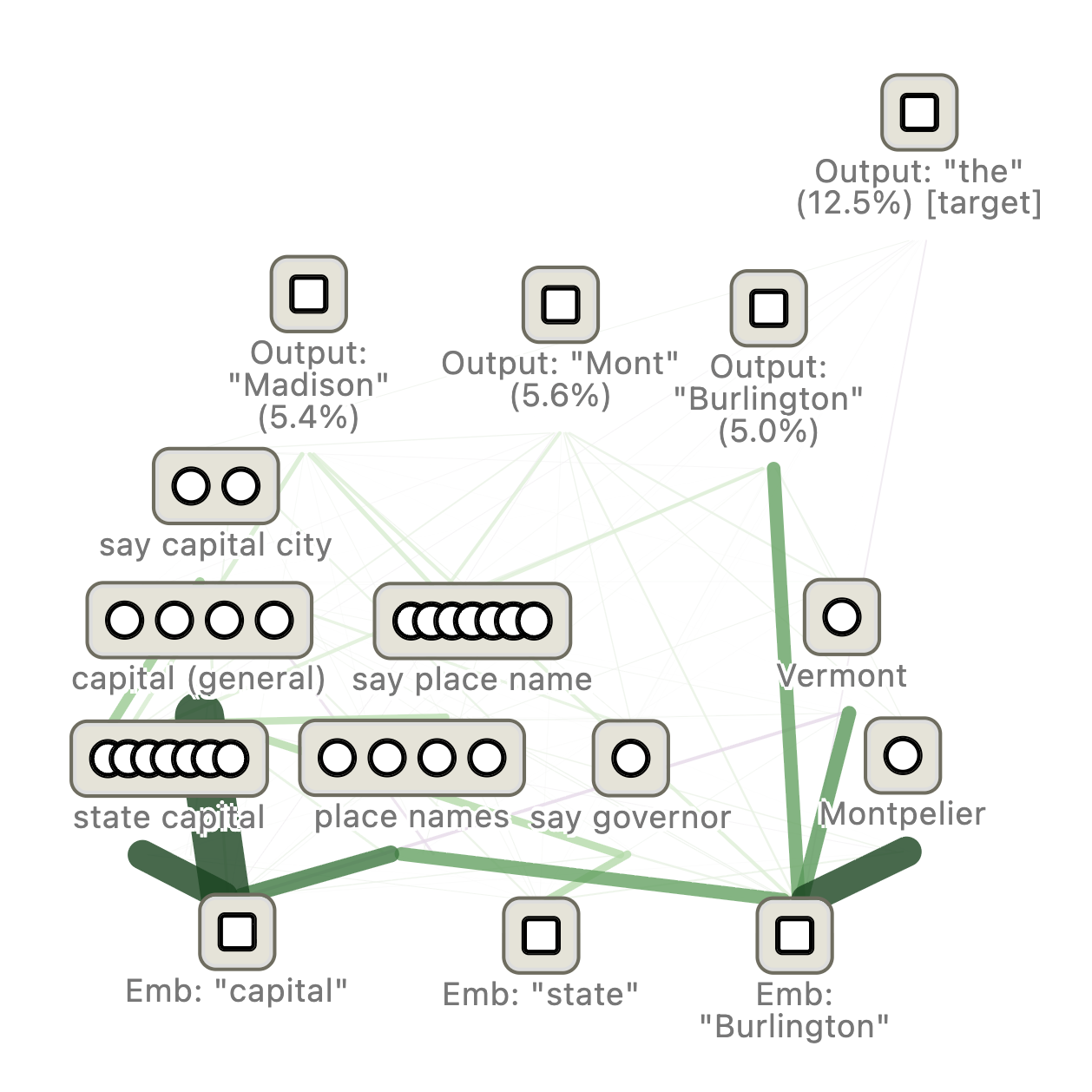}\hfill
  \includegraphics[width=0.48\textwidth]{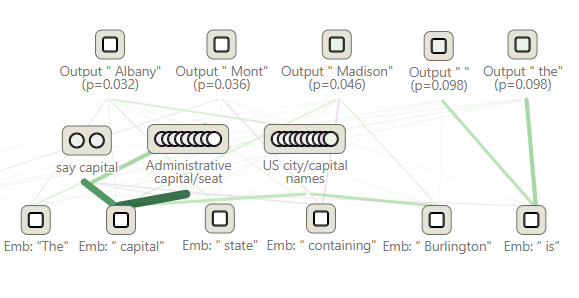}
  \caption{\textbf{Intermediate-hop supernode missing for MLP.} ``The capital of the state containing Burlington is'' $\to$ \emph{Montpelier} (intermediate hop: Vermont). \textbf{Left:} SLT transcoders. \textbf{Right:} MLP neurons. The \emph{Vermont} supernode is missing from the MLP neurons.}
  \label{fig:mlp-slt-sidebyside}
\end{figure*}

\begin{figure*}[hbt!]
  \centering
  \includegraphics[width=0.48\textwidth]{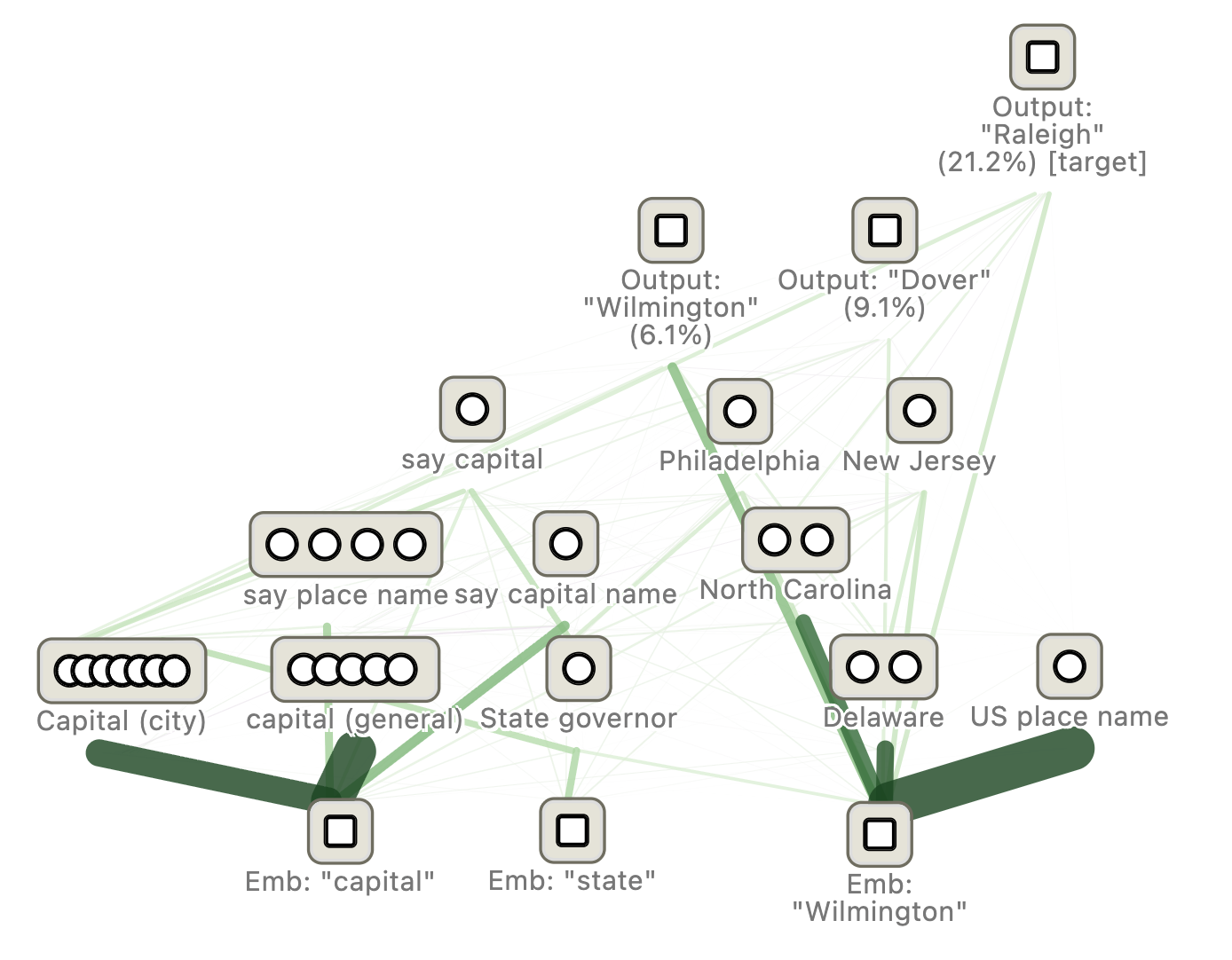}\hfill
  \includegraphics[width=0.48\textwidth]{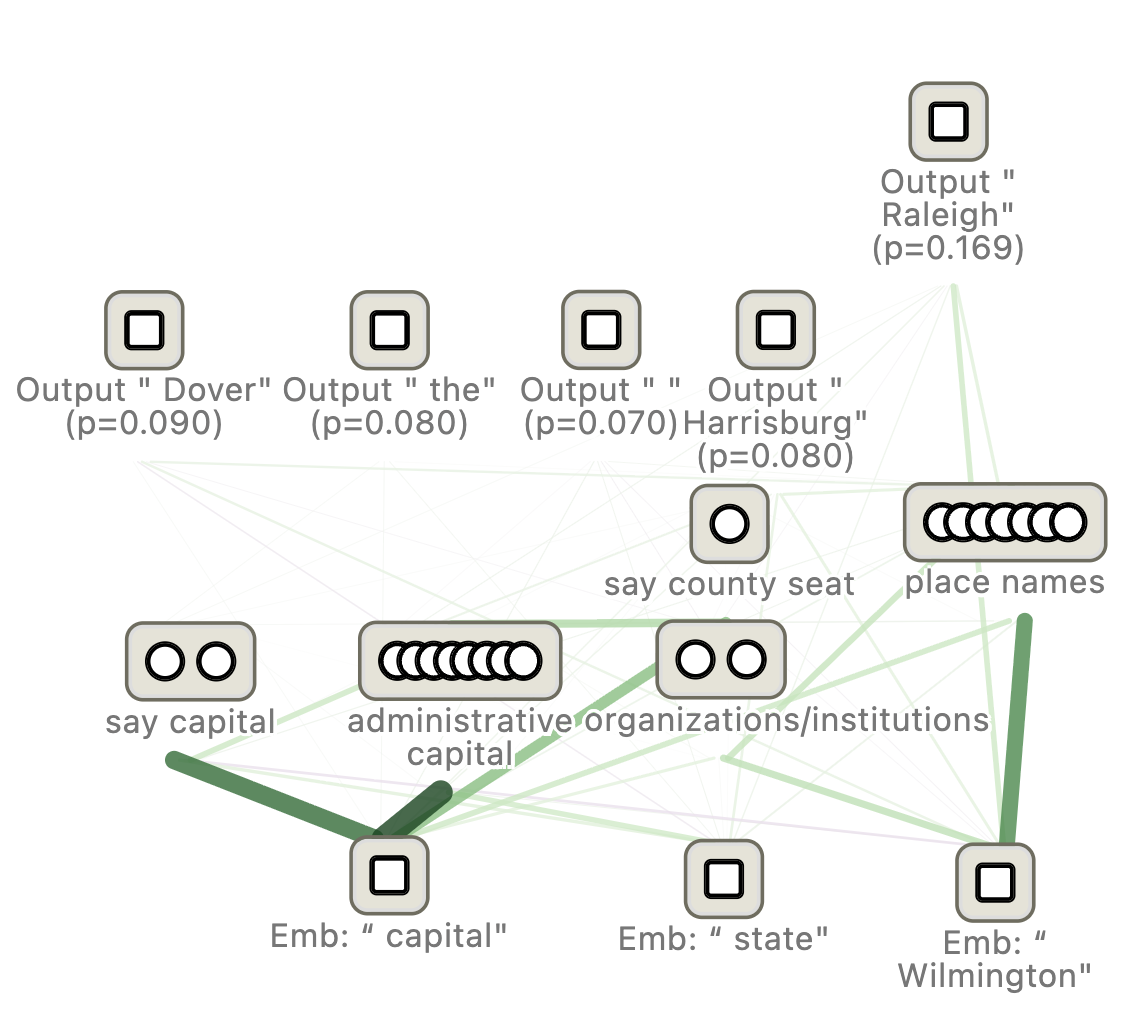}
  \caption{\textbf{Neighbor supernodes missing for MLP.} ``The capital of the state containing Wilmington is'' $\to$ \emph{Dover} (intermediate hop: Delaware). \textbf{Left:} SLT transcoders. \textbf{Right:} MLP neurons. The \emph{Philadelphia} and \emph{New Jersey} supernodes are missing from the MLP features.}
  \label{fig:mlp-slt-sidebyside2}
\end{figure*}

\end{document}